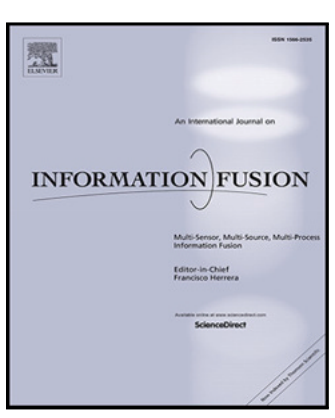

Full length article

# AI Agents vs. Agentic AI: A Conceptual taxonomy, applications and challenges

Ranjan Sapkota [a,*], Konstantinos I. Roumeliotis [b], Manoj Karkee [a,*]

[a] Cornell University, Department of Biological and Environmental Engineering, Ithaca, NY, 14850, USA
[b] University of the Peloponnese, Department of Informatics and Telecommunications, Tripoli, Greece

ABSTRACT

Information fusion, in the context of the Generative AI era, must distinguish AI Agents from Agentic AI. This review critically distinguishes between AI Agents and Agentic AI, offering a structured, conceptual taxonomy, application mapping, and analysis of opportunities and challenges to clarify their divergent design philosophies and capabilities. We begin by outlining the search strategy and foundational definitions, characterizing AI Agents as modular systems driven and enabled by LLMs and LIMs for task-specific automation. Generative AI is positioned as a precursor providing the foundation, with AI agents advancing through tool integration, prompt engineering, and reasoning enhancements. We then characterize Agentic AI systems, which, in contrast to AI Agents, represent a paradigm shift marked by multi-agent collaboration, dynamic task decomposition, persistent memory, and coordinated autonomy. Through a chronological evaluation of architectural evolution, operational mechanisms, interaction styles, and autonomy levels, we present a comparative analysis across both AI agents and agentic AI paradigms. Application domains enabled by AI Agents such as customer support, scheduling, and data summarization are then contrasted with Agentic AI deployments in research automation, robotic coordination, and medical decision support. We further examine unique challenges in each paradigm including hallucination, brittleness, emergent behavior, and coordination failure, and propose targeted solutions such as ReAct loops, retrieval-augmented generation (RAG), automation coordination layers, and causal modeling. This work aims to provide a roadmap for developing robust, scalable, and explainable AI-driven systems.

## 1. Introduction

Prior to the widespread adoption of AI Agents and Agentic AI (Fig. 1) around 2022 (Before ChatGPT was introduced), the development of autonomous and intelligent agents was deeply rooted in foundational paradigms of artificial intelligence, particularly multi-agent systems (MAS) and expert systems, which emphasized social action and distributed intelligence [1,2]. Notably, Castelfranchi [3] laid the critical groundwork by introducing ontological categories for social action, structure, and mind, arguing that sociality emerges from individual agents' actions and cognitive processes in a shared environment, with concepts like goal delegation and adoption forming the basis for cooperation and organizational behavior. Similarly, Ferber [4] provided a comprehensive framework for MAS, defining agents as entities with autonomy, perception, and communication capabilities, and highlighting their applications in distributed problem-solving, collaborative robotics, and synthetic world simulations.

These early studies established that individual social actions and cognitive architectures are fundamental to modeling collective phenomena, setting the stage for modern AI Agents. This paper builds on these foundational concepts to explore how social action modeling, as proposed in [3,4], informs the design of AI Agents capable of complex, socially intelligent interactions in dynamic environments.

Classical Agent-like systems were designed to perform specific tasks with predefined rules, which offered limited autonomy, and minimal adaptability to dynamic environments. These systems were primarily reactive or deliberative, relying on symbolic reasoning, rule-based logic, or scripted behaviors rather than the learning-driven, context-aware capabilities of modern AI Agents [5,6]. For instance, expert systems used knowledge bases and inference engines to emulate human decision-making in domains like medical diagnosis (e.g., MYCIN [7]). Other notable examples include DENDRAL [8], an expert system for molecular structure prediction; XCON [9], used for computer system configuration; and CLIPS [10], a rule-based production system framework. Systems like SOAR [11] and the subsumption architecture [12] extended symbolic and reactive logic into cognitive modeling and robotics.

* Corresponding authors.
*E-mail addresses:* rs2672@cornell.edu (R. Sapkota), mk2684@cornell.edu (M. Karkee).

In addition to task-specific reasoning, these agents supported limited forms of social interaction. Early conversational systems like ELIZA [13] and PARRY [14] simulated basic dialogue through pattern matching and script-based responses but lacked genuine understanding or contextual adaptation. Similarly, reactive agents in robotics executed sense-act cycles based on fixed control rules, as seen in early autonomous platforms like the Stanford Cart [15].

Multi-agent systems facilitated coordination among distributed entities, exemplified by auction-based resource allocation in supply chain management [16–18]. Scripted AI in video games, like NPC behaviors in early RPGs, used predefined decision trees [19]. Furthermore, BDI (Belief-Desire-Intention) architectures enabled goal-directed behavior in software agents, such as those in air traffic control simulations [20,21].

However, across these diverse systems, early AI agents shared common limitations: they lacked self-learning, generative reasoning, and adaptability to unstructured or evolving environments. These shortcomings distinguish them from Agentic AI a recent paradigm that builds on deep learning, reinforcement learning, and foundation models to enable agents with contextual awareness, continuous learning, and emergent autonomy [22].

Recent public, academic and industry interest in AI Agents and Agentic AI reflects this broader transition in system capabilities. As illustrated in Fig. 1, Google Trends data demonstrates a significant rise in global search for both terms following the emergence of large-scale generative models in late 2022. This shift is closely tied to the evolution of agent design from the pre-2022 era, where AI Agents operated in constrained, rule-based environments, to the post-LLM period marked by learning-driven, flexible/adaptive architectures [23–25]. These newer systems enable agents to refine their performance over time and interact autonomously with unstructured, dynamic inputs [26–28]. For instance, while pre-modern expert systems required manual updates to static knowledge bases, modern agents leverage emergent neural architectures to generalize across tasks [25]. The surge in trend activity reflects growing awareness of this technological leap, as researchers and practitioners seek tools that go beyond automation toward autonomy and general-purpose reasoning. Moreover, applications are no longer confined to narrow domains like simulations or logistics, but now extend to broad practical settings demanding real-time reasoning and adaptive control. This momentum, as visualized in Fig. 1, highlights the significance of recent architectural advances in scaling autonomous agents for real-world deployment.

The release of ChatGPT in November 2022 marked a pivotal inflection point in the development and public perception of artificial intelligence, catalyzing a global surge in adoption, investment, and research activity [29]. In the wake of this breakthrough, the AI landscape underwent a rapid transformation, shifting from the use of standalone LLMs toward more autonomous, task-oriented frameworks [30]. This evolution progressed through two major post-generative phases: AI Agents and Agentic AI. Initially, the widespread success of ChatGPT popularized Generative Agents, which are LLM-based systems designed to produce novel outputs such as text, images, and code from user prompts [31,32]. These agents were quickly adopted across applications ranging from conversational assistants (e.g., GitHub Copilot [33]) and content-generation platforms (e.g., Jasper [34]) to creative tools (e.g., Midjourney [35]), revolutionizing domains like digital design, marketing, and software prototyping throughout 2023 and beyond.

Although the term AI Agent was first introduced in 1998 [3], it has since evolved significantly with the rise of generative AI. Building upon this generative foundation, a new class of systems commonly referred to as AI Agents has emerged. These agents enhanced LLMs with capabilities for external tool use (e.g., API-based tools), function calling, and sequential reasoning, enabling them to retrieve real-time information and execute multi-step workflows autonomously [36,37]. Example frameworks such as AutoGPT [38] and BabyAGI (https://github.com/yoheinakajima/babyagi) highlight this transition, showcasing how LLMs could be embedded within feedback loops to dynamically plan, act, and adapt in goal-driven environments [39,40]. By late 2023, the field had advanced further into the realm of Agentic AI complex, multi-agent systems in which specialized agents collaboratively decompose goals, communicate, and coordinate toward shared objectives. In line with this evolution, Google introduced the Agent-to-Agent (A2 A) protocol in 2025 [41], a proposed standard designed to enable seamless interoperability among agents across different frameworks and vendors. The protocol is built around five core principles: embracing agentic capabilities, building on existing standards, securing interactions by default, supporting long-running tasks, and ensuring modality agnosticism. These guidelines aim to lay the groundwork for a responsive, scalable agentic infrastructure.

Architectures such as CrewAI demonstrate how these agentic frameworks can accomplish decision-making across distributed roles, facilitating intelligent behavior in high-stake applications including robotics, logistics management, and adaptive decision-support [42].

As the field progresses from Generative Agents toward increasingly autonomous systems of Agentic AI, it becomes critically important to delineate the technological and conceptual boundaries between AI Agents and Agentic AI. While both paradigms build upon LLMs and extend the capabilities of generative systems, they embody fundamentally different architectures, interaction models, and levels of autonomy. AI Agents are typically designed as single-entity systems that perform goal-directed tasks by utilizing external tools, applying sequential reasoning, and integrating real-time information to complete well-defined functions [25,43]. In contrast, Agentic AI systems are composed of multiple, specialized agents that coordinate, communicate, and dynamically allocate sub-tasks within a broader workflow to achieve a common goal(s) [22,44]. This architectural distinction highlights clear and significant differences in scalability, adaptability, and application scope.

Understanding and formalizing the taxonomy between these two paradigms **(AI Agents and Agentic AI)** is scientifically significant for several reasons. First, it enables more precise system design by aligning computational frameworks with problem complexity ensuring that AI Agents are deployed for modular, tool-assisted tasks, while Agentic AI is employed for orchestrated multi-agent operations. Moreover, it allows for appropriate benchmarking and evaluation: performance metrics, safety protocols, and resource requirements differ substantially between agents designed for carrying out individual tasks and system of distributed agents designed for accomplishing complex, coordinated tasks. Additionally, clear taxonomy reduces development inefficiencies by preventing the misapplication of design principles such as assuming inter-agent collaboration in a system designed for single-agent execution. Without this clarity, developers and practitioners risk both under-engineering complex scenarios that require agentic coordination and over-engineering simple applications that could be solved with a single AI Agent.

This article aims to clarify the differences between AI Agents and Agentic AI, providing researchers with a foundational understanding of these technologies. The objective of this study is to formalize the distinctions, establish a shared vocabulary, and provide a structured taxonomy between AI Agents and Agentic AI that informs the next generation of intelligent agent design across academic and industrial domains, as illustrated in Fig. 2.

This review also provides a comprehensive conceptual and architectural analysis of the progression from traditional AI Agents to emergent Agentic AI systems. Rather than organizing the study around formal research questions common in review articles, we adopt a sequential, layered structure that clearly lays out the historical and technical evolution of these paradigms. Beginning with a detailed description of our search strategy and selection criteria, we first establish the foundational understanding of AI Agents by analyzing their defining attributes, such as autonomy, reactivity, and tool-based execution. We then explore the critical role of foundational models, specifically LLMs and Large Image Models (LIMs), which serve as the core reasoning and perceptual engines that drive agentic behavior. Subsequent sections examine how generative AI systems have served as precursors to more dynamic, interactive agents, setting the stage for the emergence of Agentic AI. Through this perspective, we examine and present the conceptual leap from isolated, single-agent systems to orchestrated multi-agent architectures, highlighting their structural distinctions, coordination strategies, and collaborative mechanisms. We further map the architectural evolution by analyzing the core system components of both AI Agents and Agentic AI, offering comparative description of their planning, memory, orchestration, and execution layers. Building on this foundation, we review application domains spanning customer support, healthcare, research automation tasks, and robotics, while categorizing real-world deployments by system capabilities and coordination complexity. We then assess key challenges faced by both paradigms including hallucination, limited reasoning depth, causality deficits, scalability issues, and governance risks. To address these limitations, we outline opportunities for emerging solutions such as retrieval-augmented generation, tool-based reasoning, memory architectures, and simulation-based planning. The review concludes with a forward-looking roadmap that envisions the convergence of modular AI Agents and orchestrated Agentic AI in mission-critical domains such as autonomous vehicles, finance, and healthcare, and beyond. We aim to provide researchers with a structured taxonomy and actionable insights to guide the design, deployment, and evaluation of next-generation agentic AI systems.

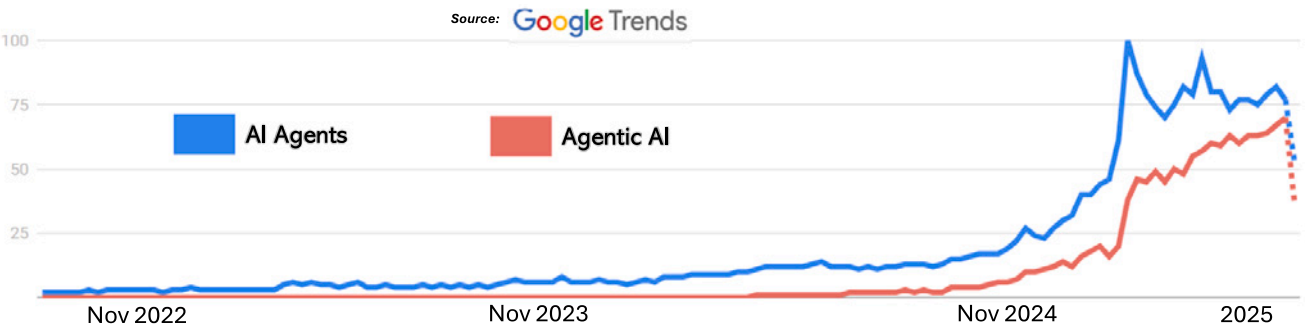

**Fig. 1.** Global Google search trends showing rising interest in “AI Agents” and “Agentic AI” since November 2022 when the ChatGPT was first introduced.

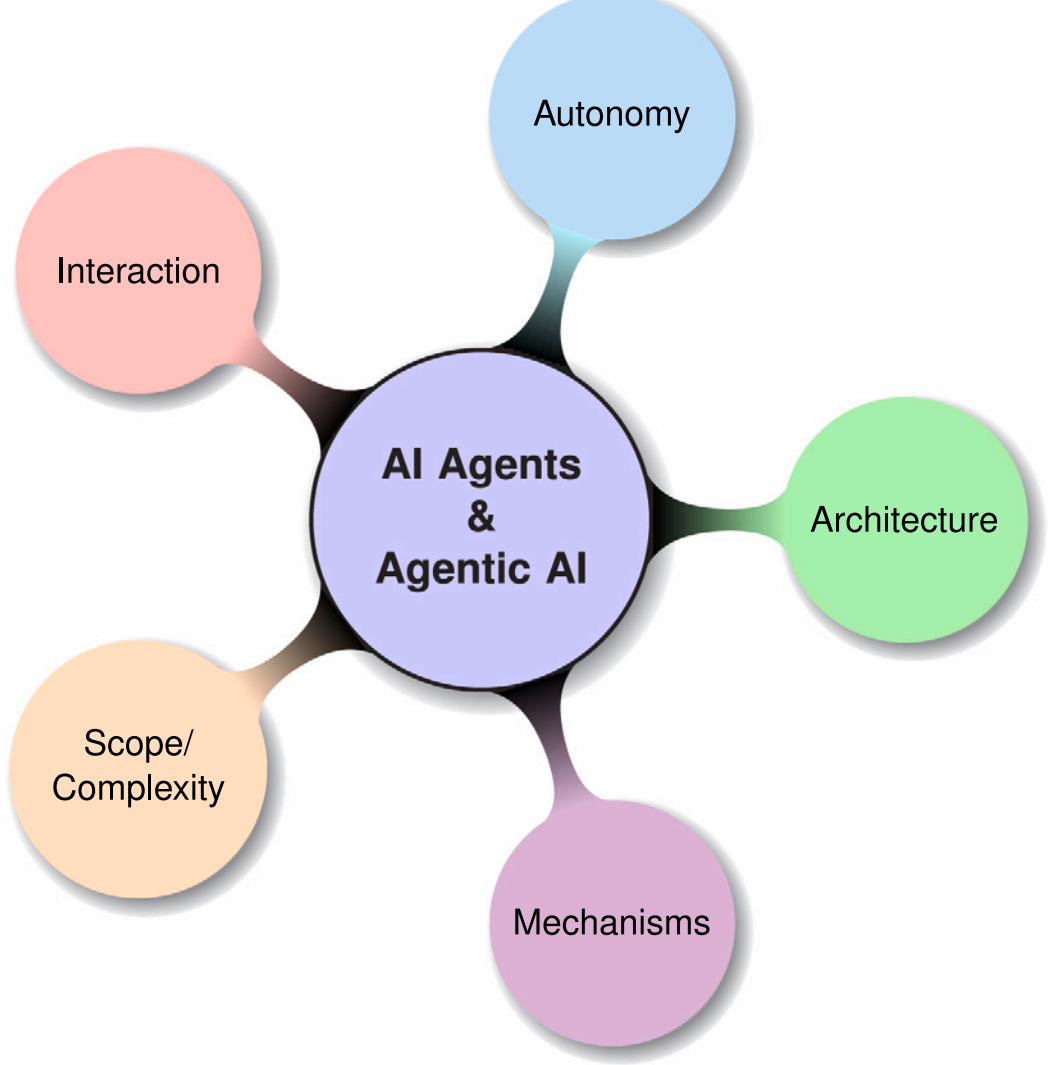

**Fig. 2.** Mind map of this research for exploring concepts, applications and challenges of AI Agents and Agentic AI. Each color-coded branch represents a key dimension of comparison: Architecture, Mechanisms, Scope/Complexity, Interaction, and Autonomy.

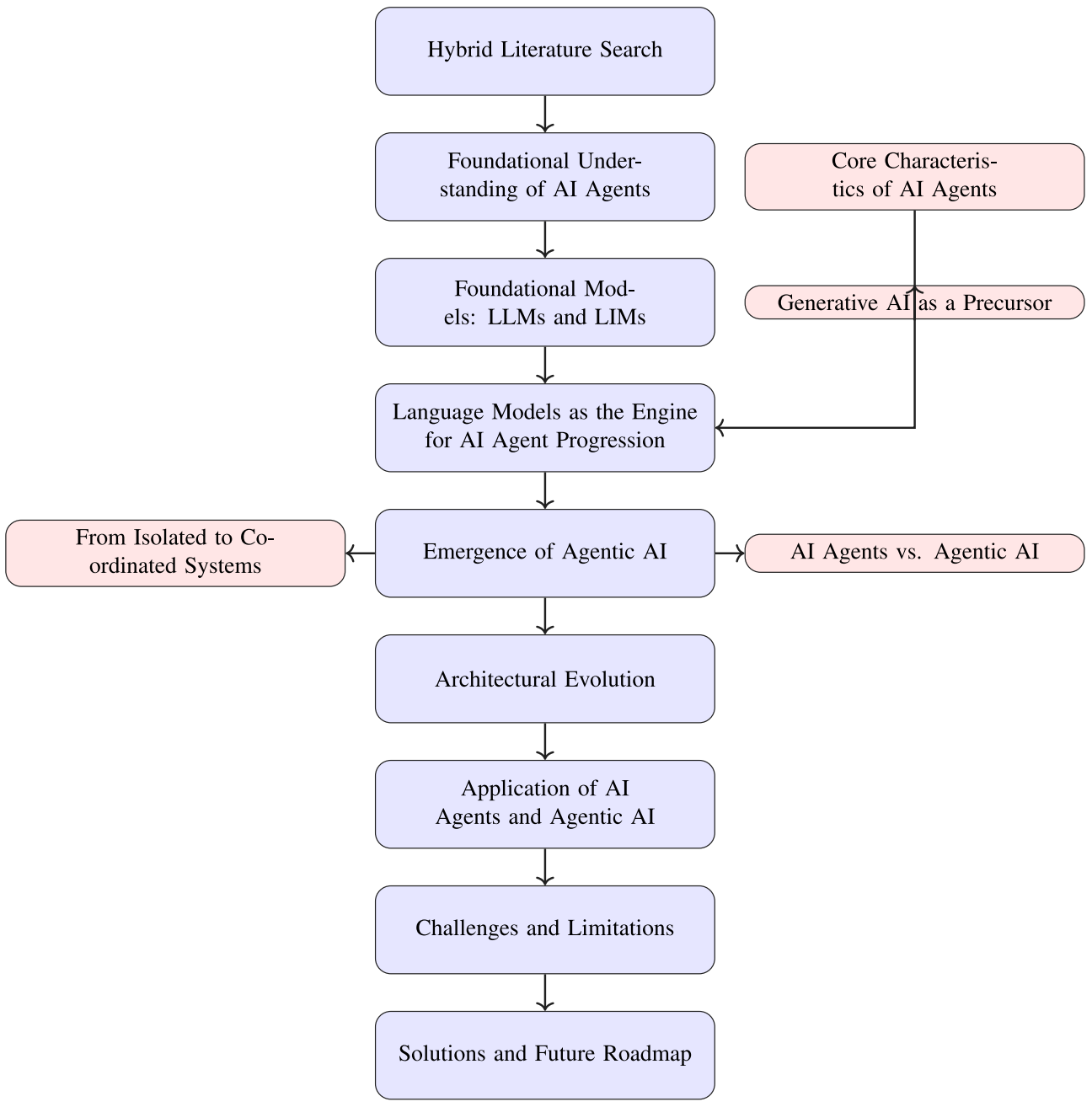

**Fig. 3.** Methodology pipeline showing the conceptual and architectural progression from foundational AI Agents to emergent Agentic AI. Each stage reflects a distinct layer of conceptual depth, highlighting key mechanisms, comparisons, challenges, and future directions.

Despite growing interest in AI Agents and Agentic AI, existing survey efforts often conflate the two under the broad umbrella of intelligent agents, leading to conceptual ambiguity and misaligned system design. Recent literature either focuses narrowly on tool-augmented LLMs for task automation or broadly discusses multi-agent systems without acknowledging the architectural and functional transition brought forth by Agentic AI. This review offers the first structured taxonomy that formally separates these paradigms across dimensions of autonomy, coordination, interaction, and reasoning scope. By positioning AI Agents as modular, single-entity systems and Agentic AI as orchestrated ecosystems with emergent behaviors, this work addresses a critical gap in distinguishing design principles, deployment strategies, and evaluation criteria. The timing of this distinction is particularly important now, as real-world applications increasingly demand scalable, multi-agent intelligence (e.g., in robotics, healthcare, and scientific discovery), yet developers often lack a clear framework for choosing between isolated agents and collaborative agentic architectures. This review bridges that gap, enabling more precise alignment of system capabilities with task complexity, and informing both research agendas and practical deployments in the post-LLM era.

### 1.1. Methodology overview

This review adopts a structured, multi-stage methodology designed to capture the evolution, architecture, application, and limitations of AI Agents and Agentic AI. The process is visually summarized in Fig. 3, which delineates the sequential flow of topics and concepts explored in this study. The analytical framework was organized to analyze and present the progression from basic agentic constructs rooted in LLMs to advanced multi-agent orchestration systems. Each step of the review was based on the rigorous synthesis of the literature from across academic sources and AI-powered platforms, enabling a comprehensive understanding of the current landscape and its emerging trends.

The review begins by establishing a *foundational understanding of AI Agents*, examining their core definitions, design principles, and architectural modules as described in the literature. These include components such as perception, reasoning, and action selection, along with early applications like customer service bots and retrieval assistants. This foundational layer serves as the conceptual entry point into the broader agentic paradigm.

Next, we discuss the role of *LLMs as core reasoning components*, emphasizing how pre-trained language models enable modern AI Agents. This section details how LLMs, through instruction fine-tuning and reinforcement learning from human feedback (RLHF), enable natural language interaction, planning, and limited decision-making capabilities. We also identify their limitations, such as hallucinations, static knowledge, and a lack of causal reasoning.

Building on these foundations, the review proceeds to the *emergence of Agentic AI*, which represents a significant conceptual advancement. Here, we highlight the transformation from tool-augmented single-agent systems to collaborative, distributed ecosystems of interacting agents. This shift is driven by the need for systems capable of decomposing goals, assigning subtasks, coordinating outputs, and adapting dynamically to changing contexts, which are the capabilities that surpass what isolated AI Agents can offer.

The next section examines the *architectural evolution from AI Agents to Agentic AI systems*, contrasting simple, modular agent designs with complex orchestration frameworks. We describe enhancements such as persistent memory, meta-agent coordination, multi-agent planning loops (e.g., ReAct and Chain-of-Thought prompting), and semantic communication protocols. Comparative architectural analysis is supported with examples from platforms like AutoGPT, CrewAI, and LangGraph.

Following the architectural exploration, the review presents an in-depth analysis of *application domains* where AI Agents and Agentic AI are being deployed. The paper discusses four representative application areas for each paradigm. For AI Agents, these include Customer Support Automation, Internal Enterprise Search, Email Filtering and Prioritization, and Personalized Content Recommendation. For Agentic AI, the applications span Multi-Agent Research Assistants, Intelligent Robotics Coordination, Collaborative Medical Decision Support, and Adaptive Workflow Automation. These use cases are examined with respect to system complexity, real-time decision-making, and collaborative task execution.

Subsequently, we address the *challenges and limitations* inherent to both paradigms. For AI Agents, we focus on issues like hallucination, prompt brittleness, limited planning ability, and lack of causal understanding. For Agentic AI, we identify higher-order challenges such as inter-agent misalignment, error propagation, unpredictability of emergent behavior, explainability deficits, and adversarial vulnerabilities. These problems are critically examined with references to recent experimental studies and technical reports.

Finally, the review outlines *potential solutions to overcome these challenges*, drawing on recent advances in causal modeling, retrieval-augmented generation (RAG), multi-agent memory frameworks, and robust evaluation pipelines. These strategies are discussed not only as technical fixes but as foundational requirements for scaling agentic systems into high-stakes domains such as healthcare, finance, and autonomous robotics.

In summary, this methodological structure enables a systematic and comprehensive assessment of the state of AI Agents and Agentic AI. By sequencing the analysis from foundational understanding, to model integration, architectural advancements, applications, and to limitations and potential solutions, the study aims to provide both theoretical clarity and practical guidance to researchers and practitioners navigating this rapidly evolving field.

#### 1.1.1. Search strategy

To develop this review, we implemented a hybrid search methodology combining traditional academic repositories and AI-enhanced literature discovery tools. Specifically, twelve platforms were queried: academic databases such as Google Scholar, IEEE Xplore, ACM Digital Library, Scopus, Web of Science, ScienceDirect, and arXiv; and AI-powered interfaces including ChatGPT, Perplexity.ai, DeepSeek, Hugging Face Search, and Grok. Search queries incorporated Boolean combinations of terms such as “AI Agents”, “Agentic AI”, “LLM Agents”, “Tool-augmented LLMs”, and “Multi-Agent AI Systems”.

Targeted queries such as “Agentic AI + Coordination + Planning”, and “AI Agents + Tool Usage + Reasoning” were also employed to retrieve papers addressing both conceptual underpinnings and system-level implementations. Literature inclusion was based on their the significance in terms of novelty, empirical evaluation, architectural contribution, and citation impact. The rising global interest in these technologies, as illustrated in Fig. 1 using Google Trends data, underscores the urgency of synthesizing this emerging knowledge space.

## 2. Foundational understanding of AI Agents

AI Agents can be defined as autonomous software entities engineered for goal-directed task execution within bounded digital environments [22,45]. These agents are defined by their ability to perceive structured or unstructured inputs [46], to reason over contextual information [47,48], and to initiate actions toward achieving specific objectives, often acting as surrogates for human users or subsystems [49]. Unlike conventional automation scripts, which follow deterministic workflows, AI Agents demonstrate reactive intelligence and some level of adaptability, allowing them to interpret dynamic inputs and reconfigure outputs accordingly [50]. Their adoption has been reported across a wide range of application domains, including customer service automation [51,52], personal productivity assistance [53], organizational information retrieval [54,55], and decision support systems [56,57].

A notable example of autonomous AI agents in Anthropic’s “Computer Use” project computer use, which showcases how their Claude model can interact with a computer in much the same way a human would. In this project, Claude is trained to visually interpret what is on a computer screen, control the mouse and keyboard, and navigate through various software applications. This allows Claude to automate repetitive tasks, such as filling out forms or copying data, as well as more complex activities like building and testing software by opening code editors, running commands, and debugging issues. Beyond these structured tasks, Claude can also handle open-ended assignments like conducting online research, gathering and organizing information from multiple sources, and even creating calendar events based on its findings. The key innovation is that Claude operates in an “agent loop”, where it receives a goal, decides on the next action, performs that action, observes the result, and repeats this process until the task is complete. This enables Claude to independently use existing computer tools and interfaces to accomplish a wide range of objectives, making it a powerful example of how autonomous AI agents can automate both routine and complex workflows.

#### 2.0.1. Core characteristics of AI Agents

AI Agents are widely conceptualized as instantiated operational instances of artificial intelligence designed to interface with users, software ecosystems, or digital infrastructures to develop goal-directed behavior [58–60]. These agents are different than general-purpose LLMs in the sense that they exhibit structured initialization, bounded autonomy, and persistent task orientation. While LLMs primarily function as reactive prompt followers [61], AI Agents operate automatically within explicitly defined scopes, engaging dynamically with inputs and producing actionable outputs in real-time environments [62].

Fig. 4 illustrates the three foundational characteristics commonly incorporated by architectural taxonomies and empirical deployments of AI Agents. These characteristics include *autonomy*, *task-specificity*, and *reactivity with adaptation*.

Together, these three characteristics provide a foundational framework for understanding and evaluating AI Agents across deployment scenarios. The remainder of this section elaborates on each characteristic, offering theoretical background and illustrative examples.

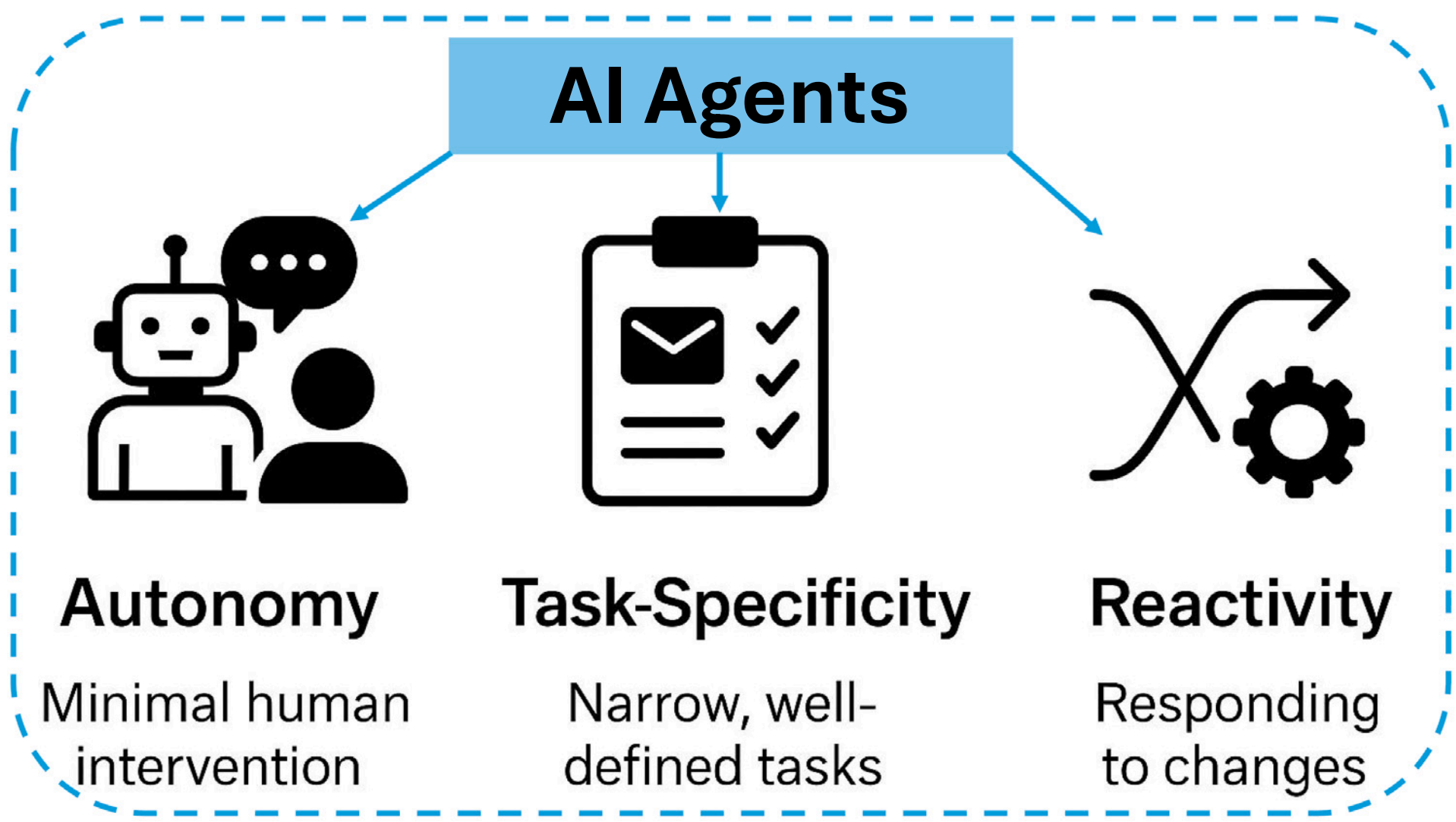

**Fig. 4.** Illustration of the three core characteristics defining AI Agents: autonomy, task-specificity, and reactivity. These traits form the foundational principles for agent design, enabling goal-oriented, adaptive behaviors within bounded environments and differentiating AI Agents from general-purpose or generative models.

- **Autonomy:** A central feature of AI Agents is their ability to function with minimal or no human intervention after deployment [63]. Once initialized, these agents are capable of perceiving environmental inputs, reasoning over contextual data, and executing predefined or adaptive actions in real-time [25]. Autonomy enables scalable deployment in applications where persistent oversight (human-in-the-loop) is impractical, such as customer support bots or scheduling assistants [52,64].
- **Task-Specificity:** AI Agents are purpose-built for narrow, and well-defined tasks [65,66]. They are optimized to execute repeatable operations within a fixed domain, such as email filtering [67,68], database querying [69], or calendar coordination [44,70]. This task specialization allows for efficiency, interpretability, and high precision in automating tasks where general-purpose reasoning is unnecessary or inefficient.
- **Reactivity and Adaptation:** AI Agents often include basic mechanisms for interacting with dynamic inputs, allowing them to respond to real-time stimuli such as user requests, external API calls, or state changes in software environments [25,71]. Some systems integrate basic learning capabilities [72] through feedback loops [73,74], heuristics [75], or updated context buffers to refine behavior over time, particularly in settings like personalized recommendations or conversation flow management [76–78].

These core characteristics collectively enable AI Agents to serve as modular, lightweight interfaces between pretrained AI models and domain-specific utility pipelines. Their architectural simplicity and operational efficiency position them as key enablers of scalable automation across enterprise, consumer, and industrial settings. Although there are currently no studies explicitly involving AI Agents integrated with specialized reasoning LLMs, their high usability and performance within constrained task boundaries have made them foundational components in contemporary intelligent system design.

#### *2.0.2. Foundational models: The role of LLMs and LIMs*

The progress in AI Agents has been significantly accelerated by the foundational development and deployment of LLMs and LIMs, which serve as the core reasoning and perception engines in contemporary agent systems. These models enable AI agents to interact intelligently with their environments, understand multi-modal inputs, and perform complex reasoning tasks that go beyond hard-coded automation.

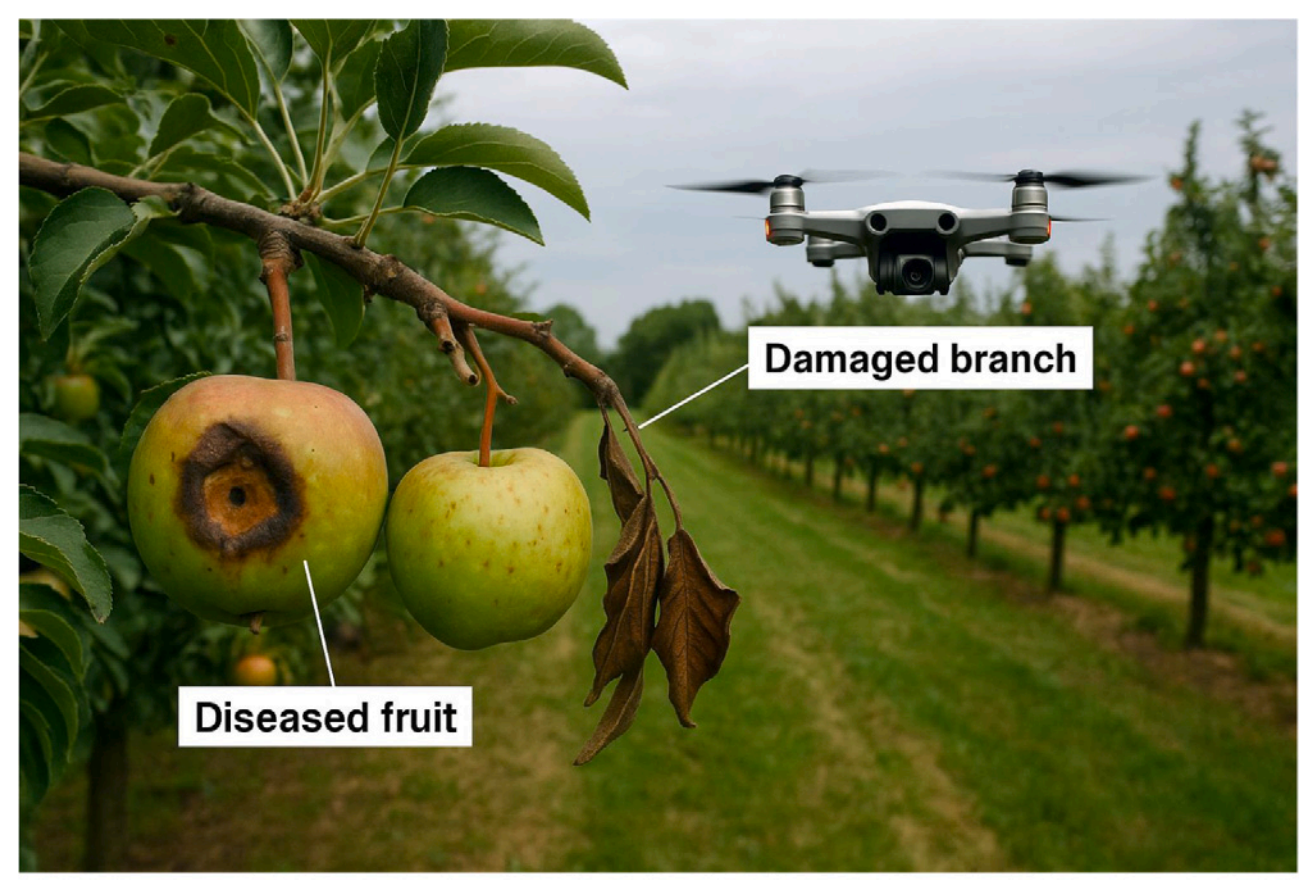

**Fig. 5.** An AI agent–enabled drone autonomously inspects an orchard, identifying diseased fruits and damaged branches using vision models, and triggers real-time alerts for targeted horticultural interventions.

**LLMs** such as GPT-4 [79] and PaLM [80] are trained on massive datasets of text from books, web content, and dialogue corpora. These models exhibit emergent capabilities in natural language understanding, question answering, summarization, dialogue coherence, and even symbolic reasoning [81–83]. Within AI Agent architectures, LLMs serve as the primary decision-making engine, allowing the agent to parse user queries, plan multi-step solutions, and generate human-like responses. For instance, an AI customer support agent powered by GPT-4 can interpret customer complaints, query backend systems via tool integration, and respond in a contextually appropriate and emotionally aware manner [84,85].

**Large Image Models (LIMs)** such as CLIP [86] and BLIP-2 [87] extend the agent's capabilities into the visual domain. Trained on image-text pairs, LIMs enable perception-based tasks including image classification, object detection, and vision-language grounding. These capabilities are increasingly vital for agents operating in domains such as robotics [88], autonomous vehicles [89,90], and visual content moderation [91,92].

For example, as illustrated in Fig. 5 where an autonomous drone agent is tasked with monitoring orchards, a LIM can identify diseased fruits [93] or damaged branches by interpreting live aerial imagery. Upon detection, the system autonomously triggers predefined intervention protocols, such as notifying horticultural staff or marking the location for targeted treatment without requiring human involvement [25, 63]. This workflow exemplifies the autonomy and reactivity of AI Agents in agricultural environments as highlighted by recent literature indicating the growing sophistication of such drone-based AI Agents. Chitra et al. [94] provides a comprehensive overview of AI algorithms foundational to embodied agents, highlighting the integration of computer vision, SLAM, reinforcement learning, and sensor fusion. These components collectively support real-time perception and adaptive navigation in dynamic environments. Kourav et al. [95] further emphasize the role of natural language processing and LLMs in generating drone action plans from human-issued queries, demonstrating how LLMs support naturalistic interaction and mission planning. Similarly, Natarajan et al. [96] explore deep learning and reinforcement learning for scene understanding, spatial mapping, and multi-agent coordination in aerial robotics. These studies converge on the critical importance of AI-driven autonomy, perception, and decision-making in advancing drone-based agents.

Importantly, LLMs and LIMs are often accessed via inference APIs provided by cloud-based platforms such as OpenAI Open AI, HuggingFace Hugging Face, and Google Gemini Google Gemini. These services abstract away the complexity of model training and fine-tuning, enabling developers to rapidly build and deploy agents equipped with state-of-the-art reasoning and perceptual abilities. This integrability accelerates prototyping and allows agent frameworks like LangChain [97] and AutoGen [98] to orchestrate LLM and LIM outputs across task workflows. In short, foundational AI models give modern AI Agents their basic understanding of language and scenes. Language models help them reason with words, and image models help them understand pictures; working together, they allow AI Agents to make smart decisions in complex situations.

#### *2.0.3. Generative AI as a precursor*

A consistent theme in the literature is the positioning of generative AI as the foundational precursor to agentic intelligence. These systems primarily operate on pre-trained LLMs and LIMs, which are optimized to synthesize multi-modal content including text, images, audio, or code based on input prompts. While highly communicative, generative models fundamentally exhibit reactive behavior: they produce output only when explicitly prompted and do not pursue goals autonomously or engage in self-initiated reasoning [99,100].

**Key Characteristics of Generative AI:**

- **Reactivity:** As non-autonomous systems, generative models are exclusively input-driven [101,102]. Their operations are triggered by user-specified prompts and they lack internal states, persistent memory, or goal-following mechanisms [103–105].
- **Multi-modal Capability:** Modern generative systems can produce a diverse array of outputs, including coherent narratives, executable code, realistic images, and even speech transcripts. For instance, models like GPT-4 [79], PaLM-E [106], and BLIP-2 [87] demonstrate these capabilities, enabling language-to-image, image-to-text, and cross-modal synthesis tasks.
- **Prompt Dependency and Statelessness:** Although generative systems are stateless in that they do not retain context across interactions unless explicitly prompted [107,108], recent advancements like GPT-4.1 support larger context windows-up to 1 million tokens-and are better able to utilize that context enabled by the improved long-text comprehension [109]. Their design also lacks intrinsic feedback loops [110], state management [111,112], or multi-step planning a requirement for autonomous decision-making and iterative goal refinement [113, 114].

Despite their remarkable generative fidelity, these systems are constrained by their inability to act upon the environment or manipulate digital tools independently. For instance, they cannot search the internet, parse real-time data, or interact with APIs without human-engineered wrappers or scaffolding layers. As such, they fall short of being classified as true AI Agents, whose architectures integrate perception, decision-making, and external tool-use within closed feedback loops.

The limitations of generative AI in handling dynamic tasks, maintaining state continuity, or executing multi-step plans led to the development of tool-augmented systems, commonly referred to as AI Agents [115]. These systems build upon the language processing backbone of LLMs but introduce additional infrastructure such as memory buffers, tool-calling APIs, reasoning chains, and planning routines to bridge the gap between passive response generation and active task completion. This architectural evolution marks a critical shift in AI system design: from content creation to autonomous task execution [116,117]. The trend from generative systems to AI Agents highlights a progressive layering of functionality that ultimately supports the emergence of agentic behaviors.

### *2.1. Language models as the engine for AI Agent progression*

The emergence of AI Agent as a transformative paradigm in artificial intelligence is closely tied to the evolution and repurposing of large-scale language models such as GPT-3 [118], Llama [119], T5 [120], Baichuan 2 [121] and GPT3mix [122]. A substantial and growing body of research shows that the advancement, from reactive generative models to autonomous, goal-directed agents is driven by the integration of LLMs as core reasoning engines within dynamic agentic systems. These models, originally trained for natural language processing tasks, are increasingly embedded in frameworks that require adaptive planning [123,124], real-time decision-making [125, 126], and environment-aware behavior [127].

#### *2.1.1. LLMs as core reasoning components*

LLMs such as GPT-4 [79], PaLM [80], Claude Claude 3.5 Sonnet, and LLaMA [119] are pre-trained on massive text corpora using self-supervised objectives and fine-tuned using techniques such as Supervised Fine-Tuning (SFT) and Reinforcement Learning from Human Feedback (RLHF) [128,129]. These models encode rich statistical and semantic knowledge, allowing them to perform tasks like inference, summarization, code generation, and dialogue management. However, in agentic contexts, their capabilities extend beyond response generation. They function as cognitive engines that interpret user goals, formulate and evaluate possible action plans, select the most appropriate strategies, leverage external tools, and manage complex, multi-step workflows.

Recent work identifies these models as central to the architecture of contemporary agentic systems. For instance, AutoGPT [38] and BabyAGI BabyAGI use GPT-4 as both a planner and executor: the model analyzes high-level objectives, decomposes them into actionable subtasks, invokes external APIs as needed, and monitors progress to determine subsequent actions. In such systems, the LLM operates in a loop of prompt processing, state updating, and feedback-based correction, closely emulating autonomous decision-making.

#### *2.1.2. Tool-augmented AI Agents: Enhancing functionality*

To overcome limitations inherent to generative-only systems such as hallucination, static knowledge cutoffs, and restricted interaction scopes, researchers have proposed the concept of tool-augmented AI Agents [130] such as Easytool [131], Gentopia [132], and ToolFive [133]. These systems integrate external tools, APIs, and computation platforms into the agent's reasoning pipeline, allowing for real-time information access, code execution, and interaction with dynamic data environments.

**Tool Invocation.** When an agent identifies a need that cannot be addressed through its internal knowledge such as querying a current stock price, retrieving up-to-date weather information, or executing a script, it generates a structured function call or API request [134,135]. These calls are typically formatted in JSON, SQL, or Python dictionary, depending on the target service, and routed through an orchestration layer that executes the task.

**Result Integration.** Once a response is received from the tool, the output is parsed and reincorporated into the LLM's context window. This enables the agent to synthesize new reasoning paths, update its task status, and decide on the next step. The ReAct framework [136] exemplifies this architecture by combining reasoning (Chain-of-Thought prompting) and action (tool use), with LLMs alternating between internal cognition and external environment interaction. A prominent example of a tool-augmented AI agent is ChatGPT, which, when unable to answer a query directly, autonomously invokes the Web Search API to retrieve more recent and relevant information, performs reasoning over the retrieved content, and formulates a response based on its understanding [137].

#### 2.1.3. Illustrative examples and emerging capabilities

Tool-augmented LLM-powered AI Agents have demonstrated potentials across a range of applications. In AutoGPT [38], the agent may plan a product market analysis by sequentially querying the web, compiling competitor data, summarizing insights, and generating a report. In a coding context, tools like GPT-Engineer combine LLM-driven design with local code execution environments to iteratively develop software artifacts as output produced during the development process, including source code, .exe files, documentation and configurations [138,139]. In research domains, systems like Paper-QA [140] utilize LLMs to query vectorized academic databases, grounding answers in retrieved scientific literature to ensure factual integrity.

These capabilities have opened pathways for more robust behavior of AI Agents such as long-horizon planning, cross-tool coordination, and adaptive learning loops. Nevertheless, the inclusion of tools also introduces new challenges in coordination complexity, error propagation, and context window limitations, which are all active areas of research. The progression toward AI Agents is inseparable from the strategic integration of LLMs as reasoning engines and their augmentation through structured utilization of external tools like search engines and APIs. This synergy transforms static language models into dynamic cognitive agents capable of perceiving, planning, acting, and adapting, thus setting the stage for multi-agent collaboration, persistent memory, and scalable autonomy, the characteristics of the Agentic AI systems.

As an example, Fig. 6 illustrates a representative use-case: a news query agent that performs real-time web search, summarizes retrieved documents, and generates an articulate, context-aware answer. Such workflows have been demonstrated in implementations using LangChain, AutoGPT, and OpenAI function-calling architectures.

## 3. The emergence of Agentic AI from AI Agent foundations

While AI Agents represent a significant leap in artificial intelligence capabilities, particularly in automating narrow tasks through tool-augmented reasoning, recent literature identifies notable limitations that constrain their scalability in complex, dynamic, multi-step, and/or cooperative scenarios [141–143]. These constraints have catalyzed the development of a more advanced paradigm: *Agentic AI*. This emerging class of systems extends the capabilities of traditional AI Agents by enabling multiple intelligent entities to collaboratively pursue goals through structured communication [144–146], shared memory [147,148], and dynamic role assignment [22].

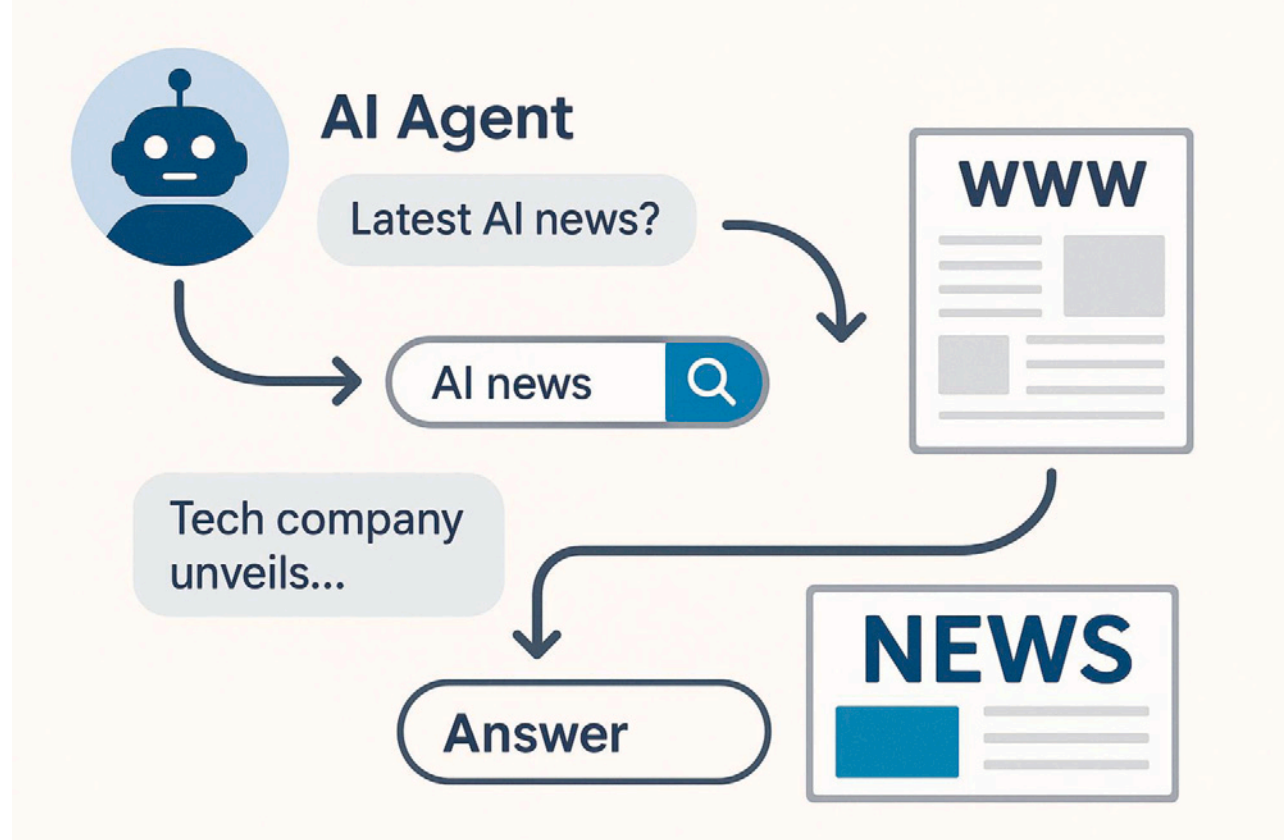

**Fig. 6.** Illustration of an AI Agent workflow for real-time news search: the agent receives a user query, performs web retrieval via external tools, summarizes retrieved articles, and generates a coherent, context-aware response. This process demonstrates autonomous tool use, reasoning, and sequential task execution.

#### 3.0.1. Conceptual leap: From isolated agents to coordinated systems

AI Agents, as explored in prior sections, integrate LLMs with external tools and APIs to execute narrowly scoped operations such as responding to customer queries, performing document retrieval, or managing schedules. However, as use cases increasingly demand context retention, task interdependence, and adaptability across dynamic environments, the single-agent model proves insufficient [149,150].

Agentic AI systems represent an emergent class of intelligent architectures in which multiple specialized agents collaborate to achieve complex, high-level objectives utilizing collaborative reasoning and multi-step planning [41]. As defined in recent frameworks, these systems are composed of modular agents each tasked with a distinct subcomponent of a broader goal and coordinated through either a centralized orchestrator or a decentralized protocol [24,145]. This structure signifies a conceptual departure from the individual, reactive behaviors typically observed in single-agent architectures, toward a form of system-level intelligence characterized by dynamic inter-agent collaboration.

A key enabler of this paradigm is **goal decomposition**, wherein a user-specified objective is automatically parsed and divided into smaller, manageable tasks by planning agents [44]. These subtasks are then distributed across the agent network. **Multi-step reasoning and planning** mechanisms facilitate the dynamic sequencing of these subtasks, allowing the system to adapt in real time to environmental changes or partial task failures. This agentic architecture ensures robust task execution even under uncertainty [22].

Inter-agent communication is mediated through **distributed communication channels**, such as asynchronous messaging queues, shared memory buffers, or intermediate output exchanges, enabling coordination without necessitating continuous central oversight [22,151]. Furthermore, **reflective reasoning and memory systems** allow agents to store context across multiple interactions, evaluate past decisions, and iteratively refine their strategies [152]. Collectively, these capabilities enable Agentic AI systems to exhibit flexible, adaptive, cooperative, and collaborative intelligence that exceeds the operational limits of individual agents.

A widely accepted conceptual illustration in the literature delineates the distinction between AI Agents and Agentic AI through the analogy of smart home systems. As depicted in Fig. 7, the left side represents a traditional AI Agent in the form of a smart thermostat. This standalone agent receives a user-defined temperature setting and autonomously controls the heating or cooling system to maintain the target temperature. While it demonstrates limited autonomy such as learning user schedules or reducing energy usage during absence, it operates in isolation, executing a singular, well-defined task without engaging in broader environmental coordination or goal inference [25,63].

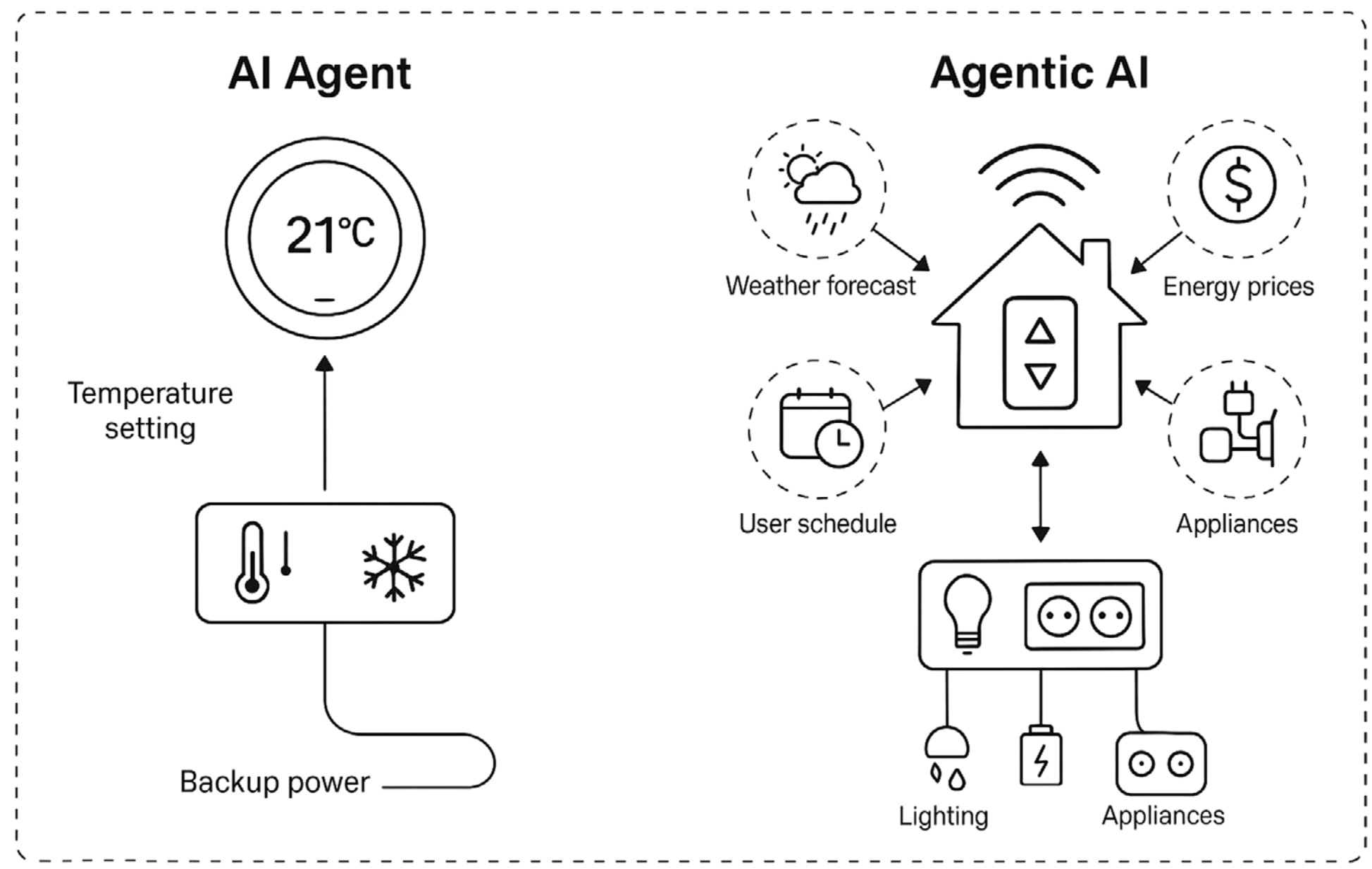

**Fig. 7.** Comparative illustration of AI Agent versus Agentic AI. The left side shows a basic AI Agent managing a single task — such as maintaining a thermostat's temperature setting based on user input. In contrast, the right side represents an Agentic AI system coordinating multiple agents to manage household automation tasks including temperature regulation, weather forecasting, energy pricing, appliance scheduling, and security monitoring. This highlights the architectural leap from narrow, reactive control to dynamic, multi-agent orchestration supporting adaptive, goal-driven intelligent environments.

In contrast, the right side of Fig. 7 illustrates an Agentic AI system embedded in a comprehensive smart home ecosystem. Here, multiple specialized agents interact synergistically to manage diverse aspects such as weather forecasting, daily scheduling, energy pricing optimization, security monitoring, and backup power activation. These agents are not just reactive modules; they communicate dynamically, share memory states, and collaboratively align actions toward a high-level system goal (e.g., optimizing comfort, safety, and energy efficiency in real-time). For instance, a weather forecast agent might signal upcoming heatwaves, prompting early pre-cooling via solar energy before peak pricing hours, as coordinated by an energy management agent. Simultaneously, the system might delay high-energy tasks or activate surveillance systems during occupant absence, integrating decisions across domains. This figure embodies the architectural and functional leap from task-specific automation to adaptive, orchestrated intelligence. The AI Agent acts as a deterministic component with limited scope, while Agentic AI reflects distributed intelligence, characterized by goal decomposition, inter-agent communication, and contextual adaptation, demonstrating key characteristics of the modern agentic AI frameworks.

### *3.0.2. Key differences between AI Agents and Agentic AI*

To systematically capture the evolution from Generative AI to AI Agents and further to Agentic AI, we structure our comparative analysis around a foundational taxonomy where Generative AI serves as the baseline. While AI Agents and Agentic AI systems represent increasingly autonomous and interactive systems, both paradigms utilize generative architectures as their foundations, especially LLMs and LIMs. Consequently, each comparative table in this subsection includes Generative AI as a reference column to highlight how agentic behavior builds on and then diverges from generative AI foundations.

A set of basic distinctions between AI Agents and Agentic AI systems, particularly in terms of scope, autonomy, architectural composition, coordination strategy, and operational complexity, are synthesized in Table 1 and Table 10 which was derived from close analysis of prominent frameworks such as AutoGen [98] and ChatDev [153]. This comparison provides a multi-dimensional view of how single-agent systems transition into coordinated, multi-agent ecosystems. Through the perspective of generative capabilities, we trace the increasing sophistication in planning, communication, and adaptation that characterizes the shift toward Agentic AI systems.

While Table 1 delineates the foundational and operational differences between AI Agents and Agentic AI systems, a more granular taxonomy is required to understand how these paradigms emerge from and relate to broader generative AI frameworks. Specifically, the conceptual and cognitive progression from static Generative AI systems to tool-augmented AI Agents, and further to collaborative Agentic AI ecosystems, necessitates an integrated comparative framework. This transition is not merely structural but also functional encompassing how initiation mechanisms, memory use, learning capacities, and orchestration strategies evolve across the agentic spectrum. Moreover, recent studies suggest the emergence of hybrid paradigms such as "Generative Agents", which blend generative modeling with modular task specialization, further complicating the agentic AI landscape. In order to capture these nuanced relationships, Table 2 synthesizes the key conceptual and cognitive dimensions across four archetypes: Generative AI, AI Agents, Agentic AI systems, and inferred Generative Agents. By positioning Generative AI as a baseline technology, this taxonomy highlights the scientific, structural and application continuum that spans from passive content generation to interactive task execution and finally to autonomous, multi-agent orchestration. This multi-tiered perspective is critical for understanding both the current capabilities and future trends of agentic intelligence across theoretical and applied domains.

To further operationalize the distinctions outlined in Table 1, Tables 2 and 3 extend the comparison between agent paradigms to encompass a broader spectrum of paradigms including AI agents and agentic AI. Table 3 presents key architectural and behavioral attributes that highlight how each paradigm differs in terms of primary capabilities, planning scope, interaction style, learning dynamics, and evaluation criteria. As can be seen from the tables, AI Agents are optimized for discrete task execution with limited planning horizons and rely on supervised or rule-based learning mechanisms. In contrast,

**Table 1**
Key Structural, Functional, and Operational Differences Between AI Agents and Agentic AI Systems. This table highlights the major distinctions between traditional AI Agents and more complex Agentic AI systems. It compares their definitions, levels of autonomy, capacity for handling task complexity, collaboration styles, learning and adaptation scope, and typical application domains. The comparison illustrates the evolution from task-specific, independently operating agents to coordinated, multi-agent systems capable of managing dynamic and large-scale workflows.

| Feature | AI Agents | Agentic AI |
|---|---|---|
| Definition | Autonomous software programs that perform specific tasks. | Systems of multiple AI agents collaborating to achieve complex goals. |
| Autonomy Level | High autonomy within specific tasks. | Broad level of autonomy with the ability to manage multi-step, complex tasks and systems. |
| Task Complexity | Typically handle single, specific tasks. | Handle complex, multi-step tasks requiring coordination. |
| Collaboration | Operate independently. | Involve multi-agent information sharing, collaboration and cooperation. |
| Learning and Adaptation | Learn and adapt within their specific domain. | Learn and adapt across a wider range of tasks and environments. |
| Applications | Customer service chatbots, virtual assistants, automated workflows. | Supply chain management, business process optimization, virtual project managers. |

Agentic AI systems extend this capacity through multi-step planning, meta-learning, and inter-agent communication, positioning them for use in complex environments requiring autonomous goal setting and coordination. Generative Agents, as a more recent construct, inherit LLM-centric pretraining capabilities and excel in producing multimodal content creation, yet they lack the proactive orchestration and state-persistent behaviors seen in Agentic AI systems.

The second table (Table 3) provides a process-driven comparison across three agent categories: Generative AI, AI Agents, and Agentic AI. This framing emphasizes how functional pipelines evolve from prompt-driven single-model inference in Generative AI, to tool-augmented execution in AI Agents, and finally to orchestrated agent networks in Agentic AI systems. The structure column highlights this progression: from single LLMs to integrated tool-chains and ultimately to distributed multi-agent systems. Access to external data, a key operational requirement for real-world utility, also increases in sophistication, from absent or optional in Generative AI to modular and coordinated in Agentic AI. Collectively, these comparative views reinforce that the evolution from generative to agentic paradigms is marked not just by increasing system complexity but also by deeper integration of autonomy, memory, and decision-making across multiple levels of abstraction (see Table 4).

Furthermore, to provide a deeper multi-dimensional understanding of the evolving agentic landscape, Tables 5 through 9 extend the comparison over five critical dimensions: core function and goal alignment, architectural composition, operational mechanism, scope and complexity, and interaction-autonomy dynamics. These dimensions serve to not only reinforce the structural differences between Generative AI, AI Agents, and Agentic AI(as explained in Table 4), but also introduce an emergent category of Generative Agents representing modular agents designed for embedded subtask-level generation within broader workflows [154]. Generative Agents are distinguished by their simulated human-like behavior, achieved through tightly integrated components such as language models, memory systems, and behavior planning modules, enabling them to operate believably and autonomously within typically closed-world environments. Table 5 situates the three paradigms in terms of their overarching goals and functional intent. While Generative AI centers on prompt-driven content generation, AI Agents emphasize tool-based task execution, and Agentic AI systems orchestrate full-fledged workflows. This functional expansion is mirrored architecturally in Table 6, where the system design transitions from single-model reliance (in Generative AI) to multi-agent orchestration and shared memory utilization in Agentic AI. Table 7 then outlines how these paradigms differ in their workflow execution pathways, highlighting the rise of inter-agent coordination and hierarchical communication as key drivers of agentic behavior.

Furthermore, Table 8 explores the increasing scope and operational complexity handled by these systems ranging from isolated content generation to adaptive, multi-agent collaboration in dynamic environments. Finally, Table 9 synthesizes the varying degrees of autonomy, interaction style, and decision-making granularity across the paradigms. These tables collectively establish a rigorous framework to classify and analyze agent-based AI systems, laying the groundwork for theoretically grounded evaluation and future design of autonomous, intelligent, and collaborative agents operating at scale.

Tables 5 through 9 offer a layered comparative analysis of Generative AI, AI Agents, and Agentic AI, anchoring the taxonomy in operational and architectural traits. Table 5 highlights core distinctions: Generative AI produces reactive content; AI Agents execute tool-based tasks; Agentic AI coordinates subagents for high-level workflow execution, marking a key shift in AI autonomy.

In Table 6, the architectural distinctions are made explicit, especially in terms of system composition and control logic. Generative AI relies on a single model with no built-in capability for tool use or delegation, whereas AI Agents combine language models with auxiliary APIs and interface mechanisms to augment functionality. Agentic AI extends this further by introducing multi-agent systems where collaboration, memory persistence, and orchestration protocols are central to the system's operation. This expansion is crucial for enabling intelligent delegation, context preservation, and dynamic role assignment capabilities absent in both generative and single-agent systems. Likewise in Table 7, differences in systems functionality and operation are presented, emphasizing distinctions in execution logic and information flow. Unlike Generative AI's linear pipeline (prompt → output), AI Agents implement procedural mechanisms to incorporate tool responses mid-process. Agentic AI introduces recursive task reallocation and cross-agent messaging, thus facilitating emergent decision-making that cannot be captured by static LLM outputs alone. Table 8 further reinforces these distinctions by mapping each system's capacity to handle task diversity, temporal scale, and operational robustness. Here, Agentic AI emerges as uniquely capable of supporting high-complexity goals that demand adaptive, multi-phase reasoning and execution strategies.

Furthermore, Table 9 highlights the operational and behavioral distinctions across Generative AI, AI Agents, and Agentic AI, with a particular focus on autonomy levels, interaction styles, and inter-agent coordination. Generative AI models such as GPT-3 [118] and DALL· E DALLE.3, remain reactive generating content solely in response to prompts without maintaining a persistent state or engaging in iterative reasoning. In contrast, AI Agents such as those constructed with LangChain [97] or MetaGPT [155], exhibit a higher degree of autonomy, capable of initiating external tool invocations and adapting behaviors within bounded tasks. However, their autonomy is typically confined to isolated task execution, lacking long-term state continuity or collaborative interaction.

**Table 2**
Summary of the Conceptual and Cognitive Taxonomy of AI Agent Paradigms Across Initiation, Adaptation, and Coordination Dimensions. This table synthesizes core cognitive and operational characteristics of four AI system types, comparing how they initiate tasks, adapt to goals, maintain temporal continuity, leverage memory, and coordinate actions. It captures the spectrum from stateless, prompt-driven Generative AI to highly coordinated and adaptive Agentic AI, while situating Generative Agents as modular components with localized generative capabilities within larger systems.

| Conceptual Dimension | Generative AI | AI Agent | Agentic AI | Generative Agent (Inferred) |
|---|---|---|---|---|
| Initiation Type | Prompt-triggered by user or input | Prompt or goal-triggered with tool use | Goal-initiated or orchestrated task | Prompt or system-level trigger |
| Goal Flexibility | (None) fixed per prompt | (Low) executes specific goal | (High) decomposes and adapts goals | (Low) guided by subtask goal |
| Temporal Continuity | Stateless, single-session output | Short-term continuity within task | Persistent across workflow stages | Context-limited to subtask |
| Learning/Adaptation | Static (pretrained) | (Might in future) Tool selection strategies may evolve | (Yes) Learns from outcomes | Typically static; limited adaptation |
| Memory Use | No memory or short context window | Optional memory or tool cache | Shared episodic/task memory | Subtask-level or contextual memory |
| Coordination Strategy | None (single-step process) | Isolated task execution | Hierarchical or decentralized coordination | Receives instructions from system |
| Key Role | Content generator | Tool-based task executor | Collaborative workflow orchestrator | Subtask-level modular generator |

**Table 3**
Key Differentiating Attributes of AI Agents, Agentic AI, and Generative Agents Across Capability, Learning, and Interaction Dimensions. This table outlines critical distinctions among three AI paradigms by comparing their primary capabilities, planning horizons, learning mechanisms, interaction styles, and evaluation criteria. It highlights the progression from task-specific execution in traditional AI Agents to the autonomous and adaptive behaviors of Agentic AI, and the creative, content-focused nature of Generative Agents.

| Aspect | AI Agent | Agentic AI | Generative Agent |
|---|---|---|---|
| Primary Capability | Task execution | Autonomous goal setting | Content generation |
| Planning Horizon | Single-step | Multi-step | N/A (content only) |
| Learning Mechanism | Rule-based or supervised | Reinforcement/meta-learning | Large-scale pretraining |
| Interaction Style | Reactive | Proactive | Creative |
| Evaluation Focus | Accuracy, latency | Engagement, adaptability | Coherence, diversity |

Agentic AI systems mark a significant departure from these paradigms by introducing internal orchestration mechanisms and multi-agent collaboration frameworks. For example, platforms like AutoGen [98] and ChatDev [153] exemplify agentic coordination through task decomposition, role assignment, and recursive feedback loops. In AutoGen, one agent might serve as a planner while another retrieves information and a third synthesizes a report, each communicating through shared memory buffers and governed by an orchestrator agent that monitors dependencies and overall task progression. This structured coordination allows for more complex goal pursuit and flexible behavior in dynamic environments. Such architectures fundamentally shift the focus of intelligence from single-model-based outputs to system-level behavior, wherein agents learn, adapt, and update decisions based on evolving task states. Thus, this comparative taxonomy not only highlights increasing levels of operational independence but also illustrates how Agentic AI introduces novel paradigms of communication, memory integration, and decentralized control, paving the way for the next generation of autonomous systems with scalable, adaptive intelligence.

### 3.1. Architectural evolution: From AI Agents to Agentic AI systems

While both AI Agents and Agentic AI systems utilize modular design principles, Agentic AI significantly extends the foundational architecture to support more complex, distributed, and adaptive behaviors. As illustrated in Fig. 8, the transition begins with core subsystems' Perception, Reasoning, and Action, that define traditional AI Agents. Agentic AI enhances this foundation by integrating advanced components such as Specialized Agents, Advanced Reasoning & Planning, Persistent Memory, and Orchestration. The figure further emphasizes emergent capabilities including Multi-Agent Collaboration, System Coordination, Shared Context, and Task Decomposition, all encapsulated within a dotted boundary that signifies the shift toward proactive, decentralized, and goal-driven system architectures. As mentioned before, this progression marks a fundamental inflection point in intelligent agent design. This section synthesizes findings from empirical frameworks such as LangChain [97], AutoGPT [98], and TaskMatrix [156], highlighting this progression in architectural sophistication.

#### 3.1.1. Core architectural components of AI Agents

Foundational AI Agents are typically composed of four primary subsystems: perception, reasoning, action, and learning. These subsystems form a closed-loop operational cycle, commonly referred to as "Understand, Think, Act, Learn" from a user interface perspective, or "Input, Processing, Action, Learning" in systems design literature [22,157].

- **Perception Module:** This subsystem intakes input signals from users (e.g., natural language prompts) or external systems (e.g., APIs, file uploads, sensor streams), and performs data pre-processing to create datasets in formats interpretable by the agent's reasoning module. For example, in LangChain-based agents [97, 158], the perception layer handles prompt templating, contextual wrapping, and retrieval augmentation via document chunking and embedding search.
- **Knowledge Representation and Reasoning (KRR) Module:** At the core of the agent's intelligence lies the KRR module, which applies symbolic, statistical, or hybrid logic to input data. Techniques include rule-based logic (e.g., if-then decision trees), deterministic workflow engines, or simple planning graphs. Reasoning in agents like AutoGPT [38] is enhanced with function-calling and prompt chaining to simulate thought processes (e.g., "step-by-step" prompts or intermediate tool invocations).

**Table 4**
Comparison of Generative AI, AI Agents, Agentic AI and Inferred Generative Agents Based on Core Function and Primary Goal. This table highlights the foundational purpose and operational focus of each system type, distinguishing their roles in AI workflows. It contrasts their core functions such as content generation, task execution, or workflow orchestration and clarifies the primary goals each category is optimized to achieve, from generating media to autonomously managing complex tasks.

| Feature | Generative AI | AI Agent | Generative Agent | Agentic AI |
|---|---|---|---|---|
| Core Function | Content generation | Task-specific execution using tools | Simulated human-like behavior | Complex workflow automation |
| Mechanism | Prompt → LLM → Output | Prompt → Tool Call → LLM → Output | Prompt → LLM + Memory/Planning → Output | Goal → Agent Orchestration → Output |
| Structure | Single model | LLM + tool(s) | LLM + memory + behavior model | Multi-agent system |
| External Data Access | None (unless added) | Via external APIs | Typically closed-world (simulated inputs) | Coordinated multi-agent access |
| Key Trait | Reactivity | Tool-use | Believability/Autonomy | Collaboration |

**Table 5**
Comparison Generative AI, AI Agents, Agentic AI, and Inferred Generative Agents Based on Core Function and Primary Goal. This table highlights the foundational purpose and operational focus of each system type, distinguishing their roles in AI workflows. It contrasts their core functions such as content generation, task execution, or workflow orchestration and clarifies the primary goals each category is optimized to achieve, from generating media to autonomously managing complex tasks.

| Feature | Generative AI | AI Agent | Agentic AI | Generative Agent (Inferred) |
|---|---|---|---|---|
| Primary Goal | Create novel content based on prompt | Execute a specific task using external tools | Automate complex workflow or achieve high-level goals | Perform a specific generative sub-task |
| Core Function | Content generation (text, image, audio, etc.) | Task execution with external interaction | Workflow orchestration and goal achievement | Sub-task content generation within a workflow |

**Table 6**
Comparison of Architectural Components Across Generative AI, AI Agents, Agentic AI, and Generative Agents. This table highlights key structural elements that define each AI paradigm, including core processing engines, prompt usage, tool and API integration, presence of multiple agents, and orchestration mechanisms. It illustrates the progression from single-model generative systems to complex multi-agent orchestration, while situating Generative Agents as modular units within broader workflows.

| Component | Generative AI | AI Agent | Agentic AI | Generative Agent (Inferred) |
|---|---|---|---|---|
| Core Engine | LLM/LIM | LLM | Multiple LLMs (potentially diverse) | LLM |
| Prompts | Yes (input trigger) | Yes (task guidance) | Yes (system goal and agent tasks) | Yes (sub-task guidance) |
| Tools/APIs | No (inherently) | Yes (essential) | Yes (available to constituent agents) | Potentially (if sub-task requires) |
| Multiple Agents | No | No | Yes (essential; collaborative) | No (is an individual agent) |
| Orchestration | No | No | Yes (implicit or explicit) | No (is part of orchestration) |

**Table 7**
Comparison of Operational Mechanisms Among Generative AI, AI Agents, Agentic AI, and Generative Agents. This table details the driving forces behind each system's operation, their modes of interaction, approaches to workflow management, and patterns of information flow. It highlights the transition from reactive, single-step generation in Generative AI to coordinated multi-agent workflows in Agentic AI, with Generative Agents functioning as modular contributors within broader task sequences.

| Mechanism | Generative AI | AI Agent | Agentic AI | Generative Agent (Inferred) |
|---|---|---|---|---|
| Primary Driver | Reactivity to prompt | Tool calling for task execution | Inter-agent communication and collaboration | Reactivity to input or sub-task prompt |
| Interaction Mode | User → LLM | User → Agent → Tool | User → System → Agents | System/Agent → Agent → Output |
| Workflow Handling | Single generation step | Single task execution | Multi-step workflow coordination | Single step within workflow |
| Information Flow | Input → Output | Input → Tool → Output | Input → Agent1 → Agent2 → ... → Output | Input (from system/agent) → Output |

**Table 8**
Comparison of Task Scope and Complexity Across Generative AI, AI Agents, Agentic AI, and Generative Agents. This table examines the breadth and difficulty of tasks each AI paradigm typically handles, illustrating the shift from generating single content pieces to managing complex, multi-agent workflows.

| Aspect | Generative AI | AI Agent | Agentic AI | Generative Agent (Inferred) |
|---|---|---|---|---|
| Task Scope | Single piece of generated content | Single, specific, defined task | Complex, multi-faceted goal or workflow | Specific sub-task (often generative) |
| Complexity | Low (relative) | Medium (integrates tools) | High (multi-agent coordination) | Low to Medium (one task component) |
| Example (Video) | Chatbot | Tavily Search Agent | YouTube-to-Blog Conversion System | Title/Description/Conclusion Generator |

**Table 9**
Comparison of Interaction and Autonomy Levels Among Generative AI, AI Agents, Agentic AI, and Generative Agents. This table analyzes varying degrees of autonomy, modes of external and internal interaction, and decision-making processes across AI paradigms. It highlights the progression from prompt-dependent Generative AI to highly autonomous, multi-agent coordination in Agentic AI, with Generative Agents operating autonomously within limited sub-task scopes.

| Feature | Generative AI | AI Agent | Agentic AI | Generative Agent (Inferred) |
|---|---|---|---|---|
| Autonomy Level | Low (requires prompt) | Medium (uses tools autonomously) | High (manages entire process) | Low to Medium (executes sub-task) |
| External Interaction | None (baseline) | Via specific tools or APIs | Through multiple agents/tools | Possibly via tools (if needed) |
| Internal Interaction | N/A | N/A | High (inter-agent) | Receives input from system or agent |
| Decision Making | Pattern selection | Tool usage decisions | Goal decomposition and assignment | Best sub-task generation strategy |

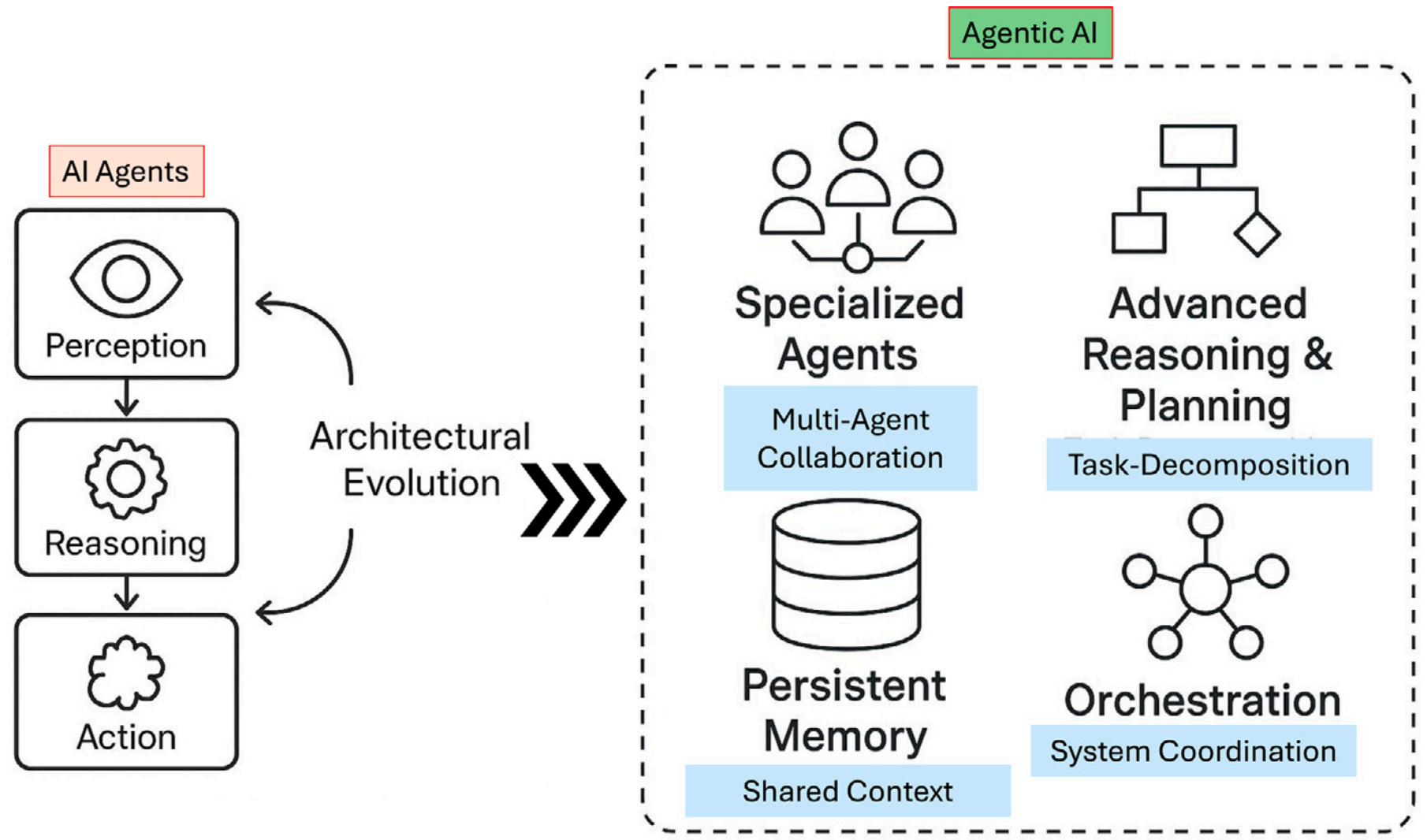

**Fig. 8.** Illustrating architectural evolution from traditional AI Agents to modern Agentic AI systems. It begins with core modules Perception, Reasoning and Action, and expands into advanced components including Specialized Agents, Advanced Reasoning & Planning, Persistent Memory, and Orchestration. The diagram further captures emergent properties such as Multi-Agent Collaboration, System Coordination, Shared Context, and Task Decomposition, all enclosed within a dotted boundary signifying layered modularity and the transition to distributed, adaptive agentic AI intelligence.

- **Action Selection and Execution Module:** This module translates inferred knowledge and decisions into external actions using an action library. These actions may include sending messages, updating databases, querying APIs, or producing structured outputs. Execution is often managed by middleware like LangChain's "agent executor", which links LLM outputs to tool calls and observes responses for subsequent steps [97].
- **Basic Learning and Adaptation:** Traditional AI Agents feature limited learning mechanisms, such as heuristic parameter adjustment [159,160] or history-informed context retention. For instance, agents may use simple memory buffers to recall prior user inputs or apply scoring mechanisms to improve tool selection in future iterations.

Customization of these agents typically involves domain-specific prompt engineering, rule injection, or workflow templates, distinguishing them from hard-coded automation scripts by their ability to make context-aware decisions. Systems like ReAct [136] exemplify this architecture, combining reasoning and action in an iterative framework where agents simulate internal dialogue before selecting external actions.

*3.1.2. Architectural enhancements in Agentic AI*

As discussed before, Agentic AI systems inherit the modularity of AI Agents but extend their architecture to support distributed intelligence, inter-agent communication, and iterative planning. The literature documents a number of critical architectural enhancements that differentiate Agentic AI from its predecessors and enable them to be highly versatile and adaptive [161,162].

- **Ensemble of Specialized Agents:** Rather than operating as a monolithic unit, Agentic AI systems consist of multiple agents, each assigned a specialized function or task (e.g., a summarizer, a retriever, or a planner). These agents interact via communication channels (e.g., message queues, blackboards, or shared memory). For instance, MetaGPT [155] highlights this approach by modeling agents after corporate departments (e.g., CEO, CTO, engineer), where roles are modular, reusable, and role-bound. In this context, "role-bound" means that each agent's behavior and responsibilities are strictly defined by its assigned role, limiting its scope of action to that specific functional domain.
- **Advanced Reasoning and Planning:** Agentic systems embed iterative reasoning capabilities using frameworks such as ReAct [136], Chain-of-Thought (CoT) prompting [163], and Tree of Thoughts [164]. These mechanisms allow agents to break down a complex task into multiple reasoning stages, evaluate intermediate results, and re-plan actions dynamically. This enables the system to respond adaptively to uncertainty or partial failure.
- **Persistent Memory Architectures:** Unlike traditional agents, Agentic AI incorporates memory subsystems to preserve and persist knowledge across task cycles or agent sessions [165,166]. Memory types include episodic memory (task-specific history) [167,168], semantic memory (long-term facts or structured data) [169,170], and vector-based memory for RAG [171,172]. For example, AutoGen [98] agents maintain scratchpads for intermediate computations, enabling stepwise task progression.
- **Orchestration Layers/Meta-Agents:** A key innovation in Agentic AI is the introduction of orchestrators meta-agents that coordinate the lifecycle of subordinate agents, manage dependencies, assign roles, and resolve conflicts. Orchestrators often include task managers, evaluators, or moderators. In ChatDev [153], for example, a virtual CEO meta-agent distributes subtasks to departmental agents and integrates their outputs into a unified strategic response.

These enhancements collectively enable Agentic AI to support scenarios that require sustained context, distributed labor, multi-modal coordination, and strategic adaptation. Use cases range from research assistants that retrieve, summarize, and draft documents in tandem (e.g., AutoGen pipelines [98]) to smart supply chain agents that monitor logistics, vendor performance, and dynamic pricing models in parallel.

The shift from isolated perception–reasoning–action loops to collaborative and self-evaluative multi-agent workflows marks a key turning point in the architectural design of intelligent systems, enabling agents not only to act but also to reflect, learn, and improve over time [173]. This progression positions Agentic AI as the next stage of AI infrastructure capable not only of executing predefined workflows but also of constructing, revising, and managing complex objectives across agents with minimal human supervision.

## 4. Application of AI Agents and Agentic AI

To illustrate the real-world utility and operational divergence between AI Agents and Agentic AI systems, this study synthesizes a range of applications drawn from recent literature, as visualized in Fig. 9. We systematically categorize and analyze application domains across two parallel tracks: conventional AI Agent systems and their more advanced Agentic AI counterparts. For AI Agents, four primary use cases are reviewed: (1) Customer Support Automation and Internal Enterprise Search, where single-agent models handle structured queries and response generation; (2) Email Filtering and Prioritization, where agents assist users in managing high-volume communication through classification heuristics; (3) Personalized Content Recommendation and Basic Data Reporting, where user behavior is analyzed for automated insights; and (4) Autonomous Scheduling Assistants, which interpret calendars and book tasks with minimal user input. In contrast, Agentic AI applications encompass broader and more dynamic capabilities, reviewed and discussed in four categories as well: (1) Multi-Agent Research Assistants that retrieve, synthesize, and draft scientific content collaboratively; (2) Intelligent Robotics Coordination, including drone and multi-robot systems in fields like agriculture and logistics; (3) Collaborative Medical Decision Support, involving diagnostic, treatment, and monitoring subsystems; and (4) Multi-Agent Game AI and Adaptive Workflow Automation, where decentralized agents interact strategically or handle complex task pipelines.

### 4.0.1. *Application of AI Agents*

1. **Customer Support Automation and Internal Enterprise Search:** AI Agents are widely adopted in enterprise environments for automating customer support and facilitating internal knowledge retrieval. In customer service, these agents leverage retrieval-augmented LLMs interfaced with APIs and organizational knowledge bases to answer user queries, triage tickets, and perform actions like order tracking or return initiation [52]. For internal enterprise search, agents built on vector stores (e.g., Pinecone, Elasticsearch) retrieve semantically relevant documents in response to natural language queries. Tools such as Salesforce Einstein Salesforce AI, Intercom Fin Fin, and Notion AI Notion AI demonstrate how structured input processing and summarization capabilities reduce workload and improve enterprise decision-making.

   A practical example (Fig. 10a) of this dual functionality can be seen in a multi-national e-commerce company deploying an AI Agent-based customer support and internal search assistant. For customer support, the AI Agent integrates with the company's Customer Relationship Management (CRM) system (e.g., Salesforce) and fulfillment APIs to resolve queries such as "Where is my order?" or "How can I return this item?". Within milliseconds, the agent retrieves contextual data from shipping databases and policy repositories, then generates a personalized response using retrieval-augmented generation. For internal enterprise search, employees use the same system to query past meeting notes, sales presentations, or legal documents. When an HR manager types "summarize key benefits of policy changes from last year", the agent queries a Pinecone vector store embedded with enterprise documentation, ranks results by semantic similarity, and returns a concise summary along with source links. These capabilities not only reduce ticket volume and support overhead but also minimize time spent searching for institutional knowledge (like policies, procedures, or manuals). The result is a unified, responsive system that enhances both external service delivery and internal operational efficiency using modular AI Agent architectures.
2. **Email Filtering and Prioritization:** As one of the important productivity tools, AI Agents automate email triage through content classification and prioritization. Integrated with systems like Microsoft Outlook and Superhuman, these agents analyze metadata and message semantics to detect urgency, extract tasks, and recommend replies. They apply user-tuned filtering rules, behavioral signals, and intent classification to reduce cognitive overload. Autonomous actions, such as auto-tagging or summarizing threads, enhance efficiency, while embedded feedback loops enable personalization through incremental learning [64]. Fig. 10b illustrates a practical implementation of AI Agents in the domain of email filtering and prioritization. In modern workplace environments, users are inundated with high volumes of email, leading to cognitive overload and missed critical communications. AI Agents embedded in platforms like Microsoft

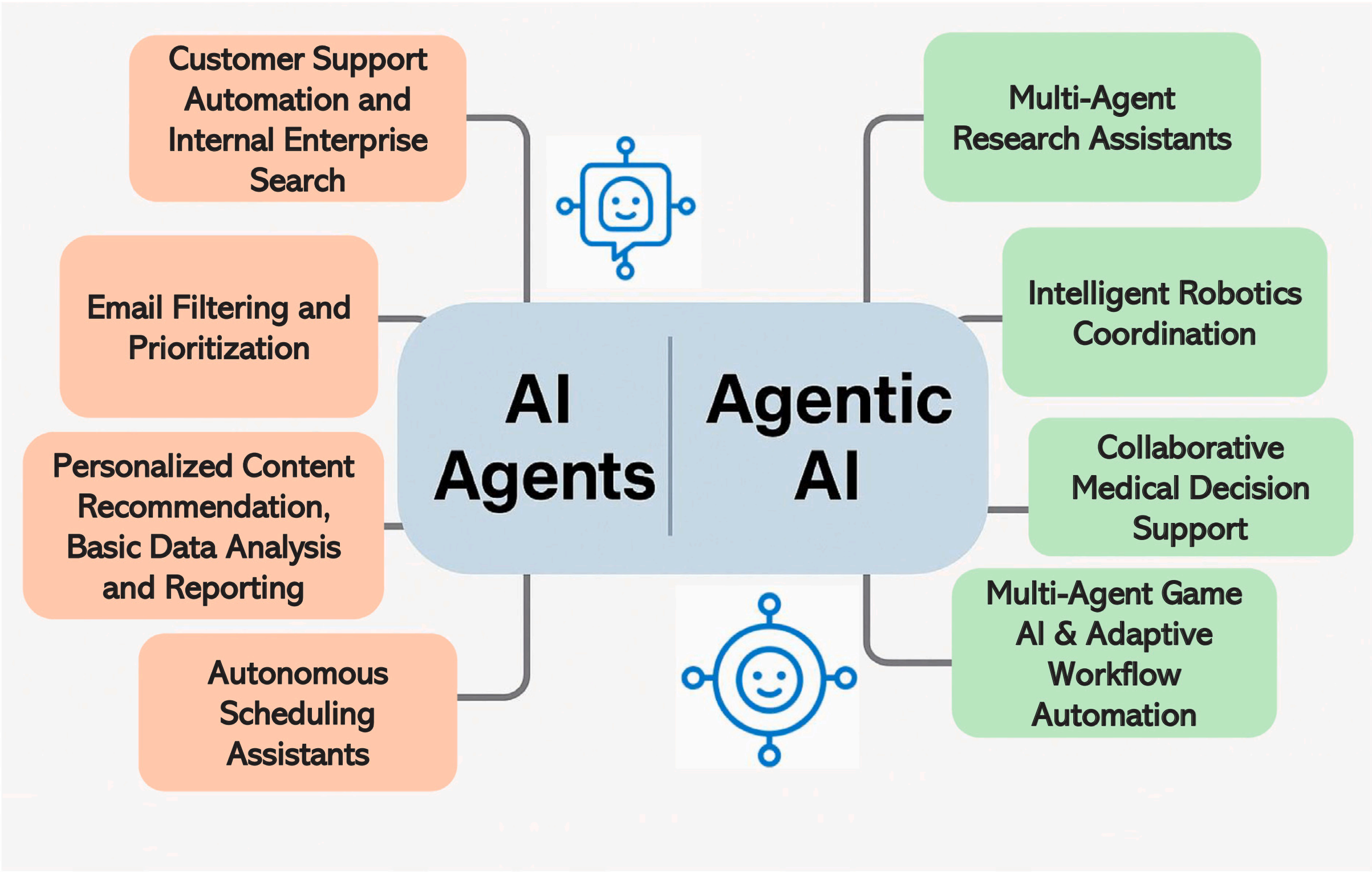

**Fig. 9.** Categorized applications of AI Agents and Agentic AI across eight core functional domains.

Outlook or Superhuman act as intelligent intermediaries that classify, cluster, and triage incoming messages. These agents evaluate metadata (e.g., sender, subject line) and semantic content to detect urgency, extract actionable items, and suggest smart replies. As depicted, the AI agent autonomously categorizes emails into tags such as "Urgent", "Follow-up", and "Low Priority", while also offering context-aware summaries and reply drafts. Through continual feedback loops and usage patterns, the system adapts to user preferences, gradually refining classification thresholds and improving prioritization accuracy. This automation offloads decision fatigue, allowing users to focus on high-value tasks, while maintaining efficient communication management in fast-paced, information-dense environments.

3. **Personalized Content Recommendation and Basic Data Reporting:** AI Agents support adaptive personalization by analyzing behavioral patterns for news, product, or media recommendations. Platforms like Amazon, YouTube, and Spotify deploy these agents to infer user preferences via collaborative filtering, intent detection, and content ranking. Simultaneously, AI Agents in analytics systems (e.g., Tableau Pulse, Power BI Copilot) enable natural-language data queries and automated report generation by converting prompts to structured database queries and visual summaries, democratizing business intelligence access.
A practical illustration (Fig. 10c) of AI Agents in personalized content recommendation and basic data reporting can be found in e-commerce and enterprise analytics systems. Consider an AI agent deployed on a retail platform like Amazon: as users browse, click, and purchase items, the agent continuously monitors interaction patterns such as dwell time, search queries, and purchase sequences. Using collaborative filtering and content-based ranking, the agent infers user intent and dynamically generates personalized product suggestions that evolve over time. For example, after purchasing gardening tools, a user may be recommended compatible soil sensors or relevant books. This level of personalization enhances customer engagement, increases conversion rates, and supports long-term user retention. Simultaneously, within a corporate setting, an AI agent integrated into Power BI Copilot allows non-technical staff to request insights using natural language, for instance, "Compare Q3 and Q4 sales in the Northeast". The agent translates the prompt into structured SQL queries, extracts patterns from the database, and outputs a concise visual summary or narrative report. This application reduces dependency on data analysts and empowers broader business decision-making through intuitive, language-driven interfaces.
4. **Autonomous Scheduling Assistants:** AI Agents integrated with calendar systems autonomously manage meeting coordination, rescheduling, and conflict resolution. Tools like x.ai and Reclaim AI interpret vague scheduling commands, access calendar APIs, and identify optimal time slots based on learned user preferences. They minimize human input while adapting to dynamic availability constraints. Their ability to interface with enterprise systems and respond to ambiguous instructions highlights the modular autonomy of contemporary scheduling agents.
A practical application of autonomous scheduling agents can be seen in corporate settings as depicted in Fig. 10d where employees manage multiple overlapping responsibilities across global time zones. Consider an executive assistant AI agent integrated with Google Calendar and Slack that interprets a command like "Find a 45 min window for a follow-up with the product team next week". The agent parses the request, checks availability for all participants, accounts for time zone differences, and avoids meeting conflicts or working-hour violations. If it identifies a conflict with a previously scheduled task, it may autonomously propose alternative windows and notify affected attendees via Slack integration. Additionally, the agent learns from historical user preferences such as avoiding early Friday meetings and refines its suggestions over time. Tools like Reclaim AI and Clockwise further illustrate this capability, offering calendar-aware automation that adapts to evolving workloads. Such assistants reduce coordination overhead, increase scheduling efficiency, and enable smoother team workflows by proactively resolving ambiguity and optimizing calendar utilization.

#### 4.0.2. Appications of Agentic AI

1. **Multi-Agent Research Assistants:** Agentic AI systems are increasingly deployed in academic and industrial research pipelines to automate multi-stage knowledge compilation. Platforms like AutoGen and CrewAI assign specialized roles to multiple agent retrievers, summarizers, synthesizers, and citation formatters under a central orchestrator. The orchestrator distributes tasks, manages role dependencies, and integrates outputs into coherent drafts or review summaries. Persistent memory allows for cross-agent context sharing and refinement over time. These systems are being used for literature reviews, grant preparation, and patent search pipelines, outperforming single-agent systems such as ChatGPT by enabling concurrent sub-task execution and long-context management [98].
For example, a real-world application of agentic AI as depicted in Fig. 11a is in the automated drafting of grant proposals. Consider a university research group preparing a National Science Foundation (NSF) submission. Using an AutoGen-based architecture, distinct agents are assigned: one retrieves prior funded proposals and extracts structural patterns; another scans recent literature to summarize related work; a third agent aligns proposal objectives with NSF solicitation language; and a formatting agent structures the document per compliance guidelines. The orchestrator coordinates these agents, resolving dependencies (e.g., aligning methodology with objectives) and ensuring stylistic consistency across sections. Persistent memory modules store evolving drafts, feedback from collaborators, and funding agency templates, enabling iterative improvement over multiple sessions. Compared to traditional manual processes, this multi-agent system significantly accelerates drafting time, improves narrative cohesion, and ensures regulatory alignment offering a scalable, adaptive approach to collaborative scientific writing in academia and R&D-intensive industries.
2. **Intelligent Robotics Coordination:** In robotics and automation, Agentic AI enable collaborative behavior in multi-robot systems. Each robot operates as a task specialized agent such as pickers, transporters, or mappers while an orchestrator supervises and adapts workflows. These architectures rely on shared spatial memory, real-time sensor fusion, and inter-agent synchronization for coordinated physical actions. Use cases include warehouse automation, drone-based orchard inspection, and robotic harvesting [155]. For instance, a swarm of agricultural drones may collectively map tree rows, identify diseased fruits, and initiate mechanical interventions. This dynamic allocation enables real-time reconfiguration and autonomy across agents facing uncertain or evolving environments.
For example, in commercial apple orchards (Fig. 11b), Agentic AI enables a coordinated multi-robot system to optimize fruit harvesting. Here, task-specialized robots such as autonomous pickers, fruit classifiers, transport bots, and drone mappers operate as agentic units under a central orchestrator. The mapping drones first survey the orchard and use vision-language models (VLMs) to generate high-resolution yield maps and identify ripe fruit clusters. This spatial data is shared via a centralized memory layer accessible by all agents. Picker robots are then assigned to high-density zones, guided by path-planning agents that optimize routes around obstacles and labor zones. Simultaneously, transport agents dynamically haul fruit containers or bins between pickers and storage, adjusting tasks in response to picker load levels and terrain changes. All agents communicate asynchronously through a shared protocol, and the coordinator continuously adjusts task priorities based on weather forecasts or mechanical faults. If one picker fails, nearby units autonomously reallocate workload. This adaptive, memory-driven coordination exemplifies Agentic AI's potential to reduce labor costs, increase harvest efficiency, and respond to uncertainties in complex agricultural environments surpassing the rigid programming of conventional agricultural robots [98,155].

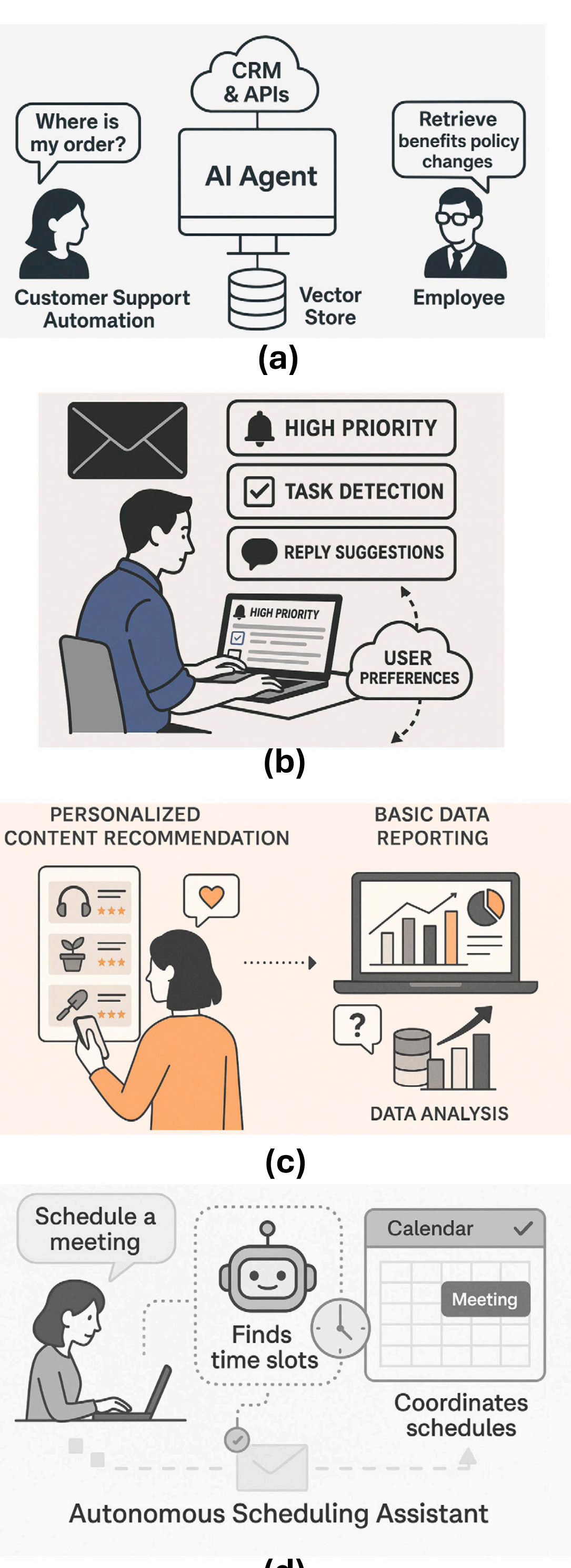

**Fig. 10.** Applications of AI Agents in enterprise settings: (a) Customer support and internal enterprise search; (b) Email filtering and prioritization; (c) Personalized content recommendation and basic data reporting; and (d) Autonomous scheduling assistants. Each example highlights modular AI Agent integration for automation, intent understanding, and adaptive reasoning across operational workflows and user-facing systems.

3. **Collaborative Medical Decision Support:** In high-stakes clinical environments, Agentic AI enables distributed medical reasoning by assigning tasks such as diagnostics, vital monitoring, and treatment planning to specialized agents. For example, one agent may retrieve patient history, another validates findings against diagnostic guidelines, and a third proposes treatment options (as seen in China's world's first Agentic AI hospital [174]). These agents synchronize through shared memory and reasoning chains, ensuring coherent and safe recommendations. Applications include ICU management, radiology triage, and pandemic response (Refer to Table 10). Although real-world implementations are still lacking due to the nascent nature of the field, studies support the potential of Agentic AI to revolutionize the healthcare sector [175].
For example, in a hospital ICU (Fig. 11c), an agentic AI system supports clinicians in managing complex patient cases. A diagnostic agent continuously analyzes vitals and lab data for early detection of sepsis risk. Simultaneously, a history retrieval agent accesses electronic health records (EHRs) to summarize comorbidities and recent procedures. A treatment planning agent cross-references current symptoms with clinical guidelines (e.g., Surviving Sepsis Campaign), proposing antibiotic regimens or fluid protocols. The orchestrator integrates these insights, ensures consistency, and surfaces conflicts for human review. Feedback from physicians is stored in a persistent memory module, allowing agents to refine their reasoning based on prior interventions and outcomes. This coordinated system enhances clinical workflow by reducing cognitive load, shortening decision times, and minimizing oversight risks. Early deployments in critical care and oncology units have demonstrated increased diagnostic precision and better adherence to evidence-based protocols, offering a scalable solution for safer, real-time collaborative medical support.
4. **Multi-Agent Game AI and Adaptive Workflow Automation:** In simulation environments and enterprise systems, Agentic AI systems facilitate decentralized task execution and effective coordination. Game platforms like AI Dungeon deploy independent NPC agents with goals, memory, and dynamic interactivity to create emergent narratives and social behavior. In enterprise workflows, systems such as MultiOn and Cognosys use agents to manage processes like legal review or incident escalation, where each step is governed by a specialized module. These architectures exhibit resilience, exception handling, and feedback-driven adaptability far beyond rule-based pipelines [176].
For example, in a modern enterprise IT environment (as depicted in Fig. 11d), Agentic AI systems are increasingly deployed to autonomously manage cybersecurity incident response workflows. When a potential threat is detected such as abnormal access patterns or unauthorized data exfiltration, specialized agents are activated in parallel. One agent performs real-time threat classification using historical breach data and anomaly detection models. A second agent queries relevant log data from network nodes and correlates patterns across systems. A third agent interprets compliance frameworks (e.g., GDPR or HIPAA) to assess the regulatory severity of the event. A fourth agent simulates mitigation strategies and forecasts operational risks. These agents coordinate under a central orchestrator that evaluates collective outputs, integrates temporal reasoning, and issues recommended actions to human analysts. Through shared memory structures and iterative feedback, the system learns from prior incidents, enabling faster and more accurate responses in future cases. Compared to traditional rule-based security systems, this agentic model enhances decision latency, reduces false positives, and supports proactive threat containment in large-scale organizational infrastructures [98] (see Tables 10 and 11).

## 5. Challenges and limitations in AI Agents and Agentic AI

To systematically understand the theoretical and operational limitations of current intelligent systems explained on Table Table 11, we present a comparative visual synthesis in Fig. 12, which categorizes challenges and potential remedies across both AI Agents and Agentic AI paradigms. Fig. 12a outlines the four most pressing limitations specific to AI Agents namely, lack of causal reasoning, inherited LLM constraints (e.g., hallucinations, shallow reasoning), incomplete agentic properties (e.g., autonomy, proactivity), and failures in long-horizon planning and recovery. These challenges often arise due to their reliance on stateless LLM prompts, limited memory, and heuristic reasoning loops.

Similarly, Fig. 12b identifies eight critical bottlenecks unique to Agentic AI systems, such as inter-agent error cascades, coordination breakdowns, emergent instability, scalability limits, and explainability issues. These challenges stem from the complexity of orchestrating multiple agents across distributed tasks without standardized architectures, robust communication protocols, or causal alignment frameworks.

Fig. 13 complements this diagnostic framework by synthesizing ten forward-looking design strategies aimed at mitigating these limitations. These include RAG, tool-based reasoning [130,131,133], agentic feedback loops (ReAct [136]), role-based multi-agent orchestration, memory architectures, causal modeling, and governance-aware design. Together, these mechanisms offer a consolidated roadmap for addressing current pitfalls and accelerating the development of safe, scalable, and context-aware autonomous systems.

#### *5.0.1. Challenges and limitations of AI Agents*

While AI Agents have garnered considerable attention for their ability to automate structured tasks using LLMs and interfaces to specific tools, the literature highlights significant theoretical and practical limitations that inhibit their reliability, generalization, and long-term autonomy [136,162]. These challenges arise from both the architectural dependence on static, pretrained models and the difficulty of instilling agentic qualities such as causal reasoning, planning, and robust adaptation. These key challenges and limitations (Fig. 12a) of AI Agents are summarized as follows:

1. **Lack of Causal Understanding:** One of the most foundational challenges lies in the agents' inability to reason causally [180,181]. While LLMs, which form the cognitive core of most AI Agents are highly effective at detecting statistical correlations within training data, they do not truly understand cause-and-effect relationships. As highlighted by recent research from DeepMind and conceptual analyses by TrueTheta [182–184], these models often fail to distinguish between mere association and actual causation. For example, an LLM-powered agent might observe that hospital visits often occur alongside illness, but it cannot determine whether the illness caused the hospital visit or vice versa. More critically, such agents cannot perform counterfactual reasoning imagining what would happen if a certain intervention or change were made. This lack of causal modeling limits their ability to make informed decisions, evaluate the impact of hypothetical actions, or provide reliable recommendations in real-world scenarios where understanding "why" something happens is essential.
Although reasoning-oriented LLMs have emerged such as DeepSeek R1 that follow a CoT approach to incrementally reason through problems, these models are not mathematically reliable reasoners (e.g., like an OWL reasoner). The chains of thought they produce are often linguistically persuasive, but not necessarily logically valid. In this sense, they do not replace formal reasoning systems such as Pellet, Bayesian networks, or causal inference frameworks, which are designed to handle logical consistency, ontological rigor, and probabilistic causality with far greater reliability.

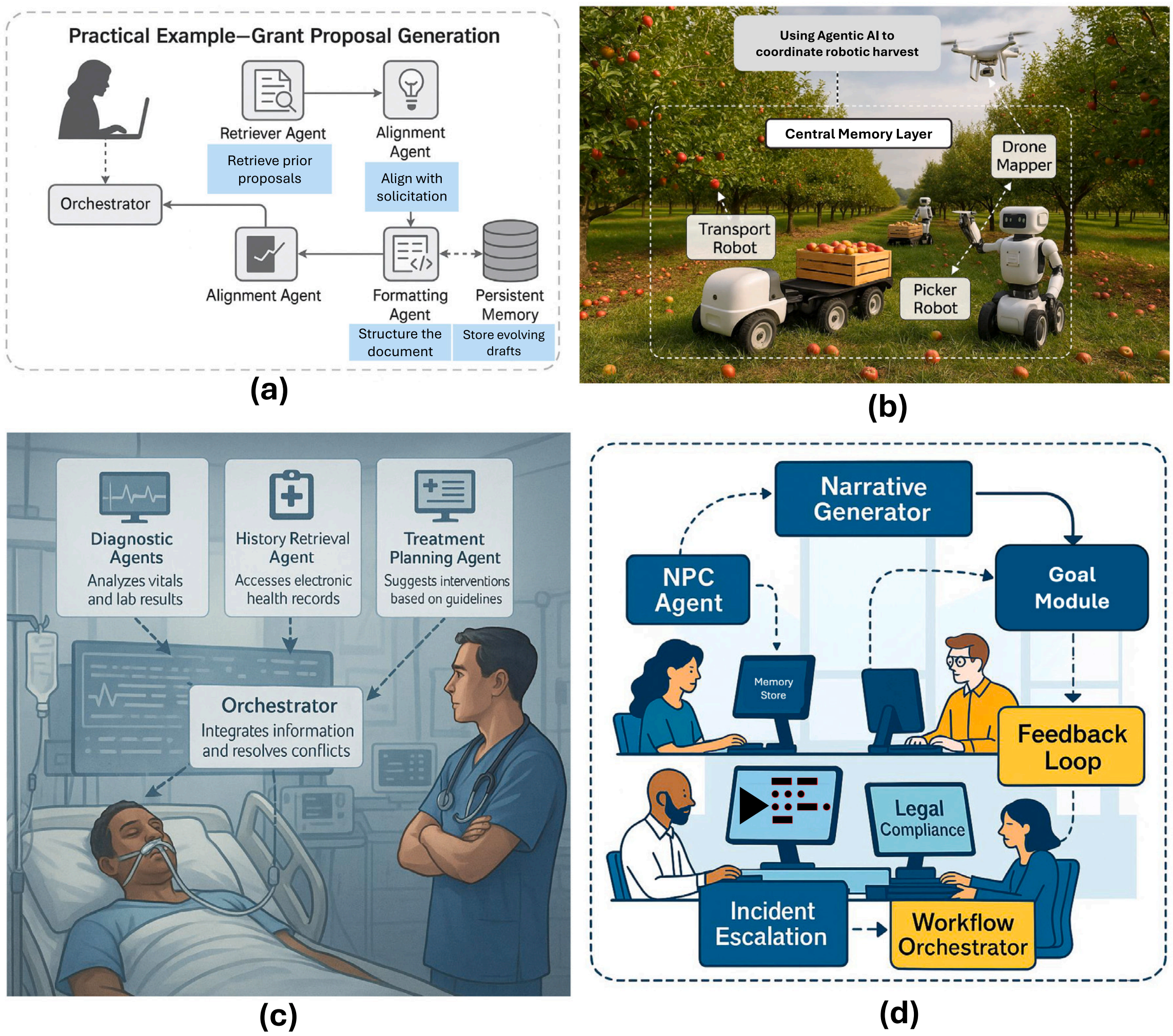

**Fig. 11. Illustrative Applications of Agentic AI Across Domains:** Fig. 11 presents four real-world applications of agentic AI systems. (a) Automated grant writing using multi-agent orchestration for structured literature analysis, compliance alignment, and document formatting. (b) Coordinated multi-robot harvesting in apple orchards using shared spatial memory and task-specific agents for mapping, picking, and transport. (c) Clinical decision support in hospital ICUs through synchronized agents for diagnostics, treatment planning, and EHR analysis, enhancing safety and workflow efficiency. (d) Cybersecurity incident response in enterprise environments via agents handling threat classification, compliance analysis, and mitigation planning. In all cases, central orchestrators manage inter-agent communication, shared memory enables context retention, and feedback mechanisms drive continual learning. These use cases highlight agentic AI's capacity for scalable, autonomous task coordination in complex, dynamic environments across science, agriculture, healthcare, and IT security.

This limitation becomes particularly challenging under distributional shifts, where real-world conditions differ from the training regime [185,186]. Without such grounding, agents remain brittle, failing in novel or high-stakes scenarios. For example, a navigation agent that excels in urban driving may misbehave in snow or construction zones if it lacks an internal causal model of road traction or spatial occlusion.

2. **Inherited Limitations from LLMs:** AI Agents, particularly those powered by LLMs, inherit a number of intrinsic limitations that impact their reliability, adaptability, and overall trustworthiness in practical deployments [187–189]. One of the most critical issues is the tendency to produce hallucinations, which are plausible but factually incorrect outputs. In high-stake domains such as legal consultation or scientific research, these hallucinations can lead to severe misjudgments and erode user trust [190,191]. Compounding this is the well-documented prompt sensitivity of LLMs, where even minor variations in phrasing can lead to divergent behaviors. This brittleness hampers reproducibility, necessitating meticulous manual prompt engineering and often requiring domain-specific tuning to maintain consistency across interactions [192].

   Furthermore, while recent agent frameworks adopt reasoning heuristics like Chain-of-Thought (CoT) [163,193] and ReAct [136] to simulate deliberative processes, these approaches remain shallow in semantic comprehension. Agents may still fail at multi-step inference, misalign task objectives, or make logically inconsistent conclusions despite the appearance of structured reasoning [136]. Such shortcomings underscore the absence of genuine understanding and generalizable planning capabilities.

   Another key limitation lies in computational cost and latency. Each cycle of agentic decision-making particularly in planning or tool-calling may require several LLM invocations. These iterations not only increase run-time latency but also scale resource consumption, creating practical bottlenecks in real-world deployments and cloud-based inference systems. Furthermore, LLMs have a static knowledge cutoff and cannot dynamically integrate new information unless explicitly augmented via retrieval or tool plugins. They also reproduce the biases of their training datasets, which can manifest as culturally insensitive or skewed responses [194,195]. Without rigorous auditing and mitigation strategies, these issues pose serious ethical and operational risks, particularly when agents are deployed in sensitive contexts or interact directly with end users (Refer to Table Table 11).

**Table 10**
Representative AI Agents (2023–2025): Applications and Operational Characteristics.

| Model/Reference | Application Area | Operation as AI Agent |
|---|---|---|
| ChatGPT Deep Research Mode<br>OpenAI (2025) Source Link | Research Analysis/Reporting | Synthesizes hundreds of sources into reports; functions as a self-directed research analyst. |
| Operator<br>OpenAI (2025) Source Link | Web Automation | Navigates websites, fills forms, and completes online tasks autonomously. |
| Agentspace: Deep Research Agent<br>Google (2025) Source Link | Enterprise Reporting | Generates business intelligence reports using Gemini models. |
| NotebookLM Plus Agent<br>Google (2025) Source Link | Knowledge Management | Summarizes, organizes, and retrieves data across Google Workspace apps. |
| Nova Act<br>Amazon (2025) Source Link | Workflow Automation | Automates browser-based tasks such as scheduling, HR requests, and email. |
| Manus Agent<br>Monica (2025) Source Link | Personal Task Automation | Executes trip planning, site building, and product comparisons via browsing. |
| Harvey<br>Harvey AI (2025) Source Link | Legal Automation | Automates document drafting, legal review, and predictive case analysis. |
| Otter Meeting Agent<br>Otter.ai (2025) Source Link | Meeting Management | Transcribes meetings and provides highlights, summaries, and action items. |
| Otter Sales Agent<br>Otter.ai (2025) Source Link | Sales Enablement | Analyzes sales calls, extracts insights, and suggests follow-ups. |
| ClickUp Brain<br>ClickUp (2025) Source Link | Project Management | Automates task tracking, updates, and project workflows. |
| Agentforce<br>Agentforce (2025) Source Link | Customer Support | Routes tickets and generates context-aware replies for support teams. |
| Microsoft Copilot<br>Microsoft (2024) Source Link | Office Productivity | Automates writing, formula generation, and summarization in Microsoft 365. |
| Project Astra<br>Google DeepMind (2025) Source Link | Multimodal Assistance | Processes text, image, audio, and video for task support and recommendations. |
| Claude 3.5 Agent<br>Anthropic (2025) Source Link | Enterprise Assistance | Uses multimodal input for reasoning, personalization, and enterprise task completion. |

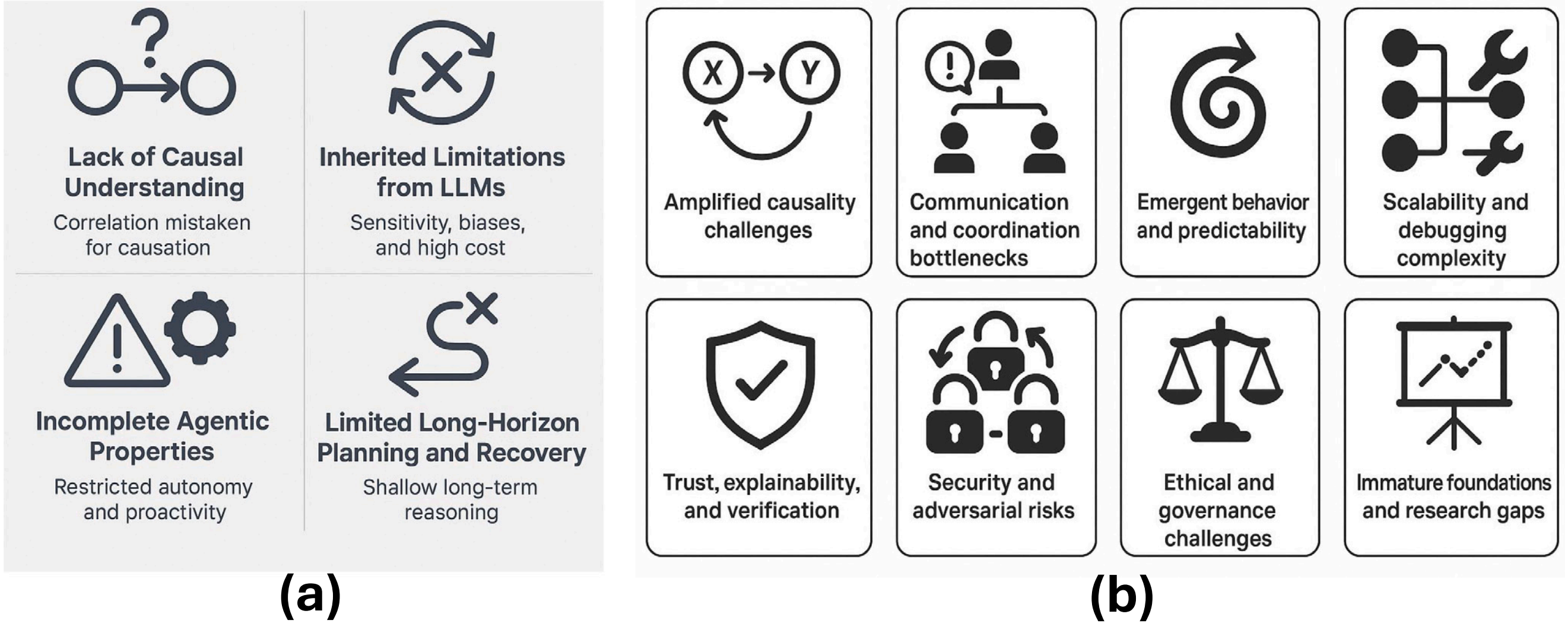

**Fig. 12.** Illustration of challenges across AI Agents and Agentic AI paradigms: (a) AI Agents face key limitations such as lack of causal reasoning, shallow planning depth, and prompt brittleness ; (b) Agentic AI systems introduce new complexities, including inter-agent misalignment, coordination failures, and emergent instability. Together, these challenges underscore the need for robust reasoning, alignment protocols, and system-level safeguards in intelligent agent design.

3. **Incomplete Agentic Properties:** A major limitation of current AI Agents is their inability to fully satisfy the canonical agentic

**Table 11**
Representative Agentic AI Models (2023–2025): Applications and Operational Characteristics.

| Model/Reference | Application Area | Operation as Agentic AI |
|---|---|---|
| Auto-GPT [38] | Task Automation | Decomposes high-level goals, executes subtasks via tools/APIs, and iteratively self-corrects. |
| GPT Engineer Open Source (2023) Source Link | Code Generation | Builds entire codebases: plans, writes, tests, and refines based on output. |
| MetaGPT [155]) | Software Collaboration | Coordinates specialized agents (e.g., coder, tester) for modular multi-role project development. |
| BabyAGI Nakajima (2024) Source Link | Project Management | Continuously creates, prioritizes, and executes subtasks to adaptively meet user goals. |
| Voyager Wang et al. (2023) [177] | Game Exploration | Learns in Minecraft, invents new skills, sets subgoals, and adapts strategy in real time. |
| CAMEL Liu et al. (2023) [178] | Multi-Agent Simulation | Simulates agent societies with communication, negotiation, and emergent collaborative behavior. |
| Einstein Copilot Salesforce (2024) Source Link | Customer Automation | Automates full support workflows, escalates issues, and improves via feedback loops. |
| Copilot Studio (Agentic Mode) Microsoft (2025) Source Link | Productivity Automation | Manages documents, meetings, and projects across Microsoft 365 with adaptive orchestration. |
| Atera AI Copilot Atera (2025) Source Link | IT Operations | Diagnoses/resolves IT issues, automates ticketing, and learns from evolving infrastructures. |
| AES Safety Audit Agent AES (2025) Source Link | Industrial Safety | Automates audits, assesses compliance, and evolves strategies to enhance safety outcomes. |
| DeepMind Gato (Agentic Mode) Reed et al. (2022) [179] | General Robotics | Performs varied tasks across modalities, dynamically learns, plans, and executes. |
| GPT-4o + Plugins OpenAI (2024) Source Link | Enterprise Automation | Manages complex workflows, integrates external tools, and executes adaptive decisions. |

properties defined in foundational literature, such as autonomy, proactivity, reactivity, and social ability [146,189]. While many systems marketed as "agents" leverage LLMs to perform useful tasks, they often fall short of these fundamental characteristics in practice. Autonomy, for instance, is typically partial at best. Although agents can execute tasks with minimal oversight once initialized, they remain heavily reliant on external scaffolding such as human-defined prompts, planning heuristics, or feedback loops to function effectively [196]. Self-initiated task generation, self-monitoring, or autonomous error correction are rare or absent, limiting their capacity for true independence.

Proactivity is similarly underdeveloped. Most AI Agents require explicit user instruction to act and lack the capacity to formulate or re-prioritize goals dynamically based on changing context/environment or evolving objectives [197]. As a result, they behave reactively rather than strategically, constrained by the static nature of their initialization. Reactivity itself is constrained by architectural bottlenecks. Agents do respond to environmental or user input, but response latency caused by repeated LLM inference calls [198,199], coupled with narrow contextual memory windows [165,200], inhibits real-time adaptability.

In addition to proactivity, social ability remains one of the most underexplored capabilities of AI Agents. Real-world AI agents and Agentic AI systems should be able to communicate and collaborate with humans or other agents over extended interactions, resolving ambiguity, negotiating tasks, and adapting to social norms. However, existing implementations exhibit brittle, template-based dialogue that lacks long-term memory integration or nuanced conversational context. Agent-to-agent interaction is often hardcoded or limited to scripted exchanges, hindering collaborative execution and emergent behavior [105,201]. Collectively, these deficiencies reveal that while AI Agents demonstrate functional intelligence, they remain far from meeting the formal benchmarks of intelligent [202], interactive, and adaptive agents. Bridging this gap is essential for advancing toward more autonomous, socially capable AI systems.

4. **Limited Long-Horizon Planning and Recovery:** A persistent limitation of current AI Agents lies in their inability to perform robust long-horizon planning, especially in complex, multi-stage tasks. This constraint stems from their foundational reliance on stateless prompt-response paradigms, where each decision is made without an intrinsic memory of prior reasoning steps unless externally managed. Although augmentations such as the ReAct framework [136] or Tree-of-Thoughts [164] introduce pseudo-recursive reasoning, they remain fundamentally heuristic and lack true internal models of time, causality, or state evolution. Consequently, agents often fail in tasks requiring extended temporal consistency or contingency planning. For example, in domains such as clinical triage or financial portfolio management, where decisions depend on prior context and dynamically varying outcomes, agents may exhibit repetitive behaviors such as endlessly querying tools or fail to adapt when subtasks fail or return ambiguous results. The absence of systematic recovery mechanisms or error detection leads to brittle workflows and error propagation. This shortfall severely limits agent deployment in mission-critical environments where reliability, fault tolerance, and sequential coherence are essential.
5. **Reliability and Safety Concerns:** AI Agents are not yet safe or verifiable enough for deployment in handling or managing critical infrastructure [203]. The absence of causal reasoning leads to unpredictable behavior under distributional shift [181,204]. Furthermore, evaluating the correctness of an agent's plan especially when the agent fabricates intermediate steps or rationales remains an unsolved problem in interpretability [114, 205]. Safety guarantees, such as formal verification, are not yet available for open-ended, LLM-powered agents. While AI Agents represent a major step beyond static generative models, their limitations in causal reasoning, adaptability, robustness, and planning restrict their deployment in high-stakes or dynamic environments. Most current systems rely on heuristic wrappers and brittle prompt engineering rather than grounded agentic cognition. Bridging this gap will require future systems to integrate causal models, dynamic memory, and verifiable reasoning mechanisms. These limitations also set the stage for the emergence of Agentic AI systems, which attempt to address these bottlenecks through multi-agent collaboration, orchestration layers, and persistent system-level context. The persistent system-level context ensures that agents operate with a shared and evolving understanding of goals, environment, and prior decisions, enabling coherent coordination and sustained autonomy across complex workflows [206,207]. This continuity is critical for reducing redundant processing and enabling long-horizon reasoning.

#### 5.0.2. Challenges and limitations of agentic AI

Agentic AI systems represent a paradigm shift from isolated AI Agents to collaborative, multi-agent ecosystems capable of decomposing and executing complex goals [22]. These systems typically consist of orchestrated or communicating agents that interact via tools, APIs, and shared environments [26,44]. While this architectural evolution enables more ambitious automation, it introduces a range of amplified and novel challenges that compound existing limitations of individual LLM-based agents. The current challenges and limitations of Agentic AI systems are as follows:

1. **Amplified Causality Challenges:** One of the most critical limitations in Agentic AI systems is the magnification of lack of causal reasoning already observed in single-agent architectures. Unlike traditional AI Agents that operate in relatively isolated environments, Agentic AI systems involve complex inter-agent dynamics and cooperation, where each agent's action can influence the decision space of others. Without a robust capacity for modeling cause-and-effect relationships, these systems struggle to coordinate effectively and adapt to unforeseen environmental shifts.
   A key issue cause by this limitation is *inter-agent distributional shift*, where the behavior of one agent alters the operational context for others. In the absence of causal reasoning, agents are unable to anticipate the downstream impact of their outputs, resulting in coordination breakdowns or redundant computations [208]. Furthermore, these systems are particularly vulnerable to *error cascades*: a faulty or hallucinated output from one agent can propagate through the system, compounding inaccuracies and corrupting subsequent decisions. For example, if a verification agent erroneously validates false information, downstream agents such as summarizers or decision-makers may unknowingly build upon that misinformation, compromising the integrity of the entire system. This fragility underscores the urgent need for integrating causal inference and intervention modeling into the design of multi-agent workflows, especially in high-stake or dynamic environments where systemic robustness is essential.
2. **Communication and Coordination Bottlenecks:** A fundamental challenge in Agentic AI lies in achieving efficient communication and coordination across multiple autonomous agents. Unlike single-agent systems, Agentic AI systems involve distributed agents that must collaboratively pursue a shared objective requiring precise alignment, synchronized execution, and robust communication protocols. However, current implementations fall short in these aspects. One major issue is *goal alignment and shared context*, where agents often lack a unified semantic understanding of overarching objectives. This lack of shared semantic grounding hampers sub-task decomposition, dependency management, and progress monitoring, especially in dynamic environments requiring causal awareness and temporal coherence.
   In addition, *protocol limitations* significantly hinder inter-agent communication. Most systems rely on natural language exchanges over loosely defined interfaces, which are prone to ambiguity, inconsistent formatting, and contextual drift. These communication gaps lead to fragmented strategies, delayed coordination, and degraded system performance. Furthermore, *resource contention* emerges as a systemic bottleneck when agents simultaneously access shared computational, memory, or API resources. Without centralized orchestration or intelligent scheduling mechanisms, these conflicts can result in race conditions, execution delays, or outright system failures. Collectively, these bottlenecks illustrate the immaturity of current coordination frameworks in Agentic AI, and highlight the pressing need for standardized communication protocols, semantic task planners, and global resource managers to ensure scalable, coherent multi-agent collaboration.
3. **Emergent Behavior and Predictability:** One of the most critical limitations of Agentic AI lies in managing emergent behaviors of complex system-level phenomena that arise from the interactions of autonomous agents. While such emergence can potentially yield adaptive and innovative solutions, it also introduces significant unpredictability and safety risks [157,209]. A key concern is the generation of *unintended outcomes*, where agent interactions result in behaviors that were not explicitly programmed or foreseen by system designers. These behaviors may diverge from task objectives, generate misleading outputs, or even lead to harmful actions particularly in high-stake domains like healthcare, finance, or critical infrastructure.
   As the number of agents and the complexity of their interactions grow, so too does the likelihood of *system instability*. This includes phenomena such as infinite planning loops, action deadlocks, and contradictory behaviors emerging from asynchronous or misaligned agent decisions. Without centralized arbitration mechanisms, conflict resolution protocols, or fallback strategies, these instabilities compound over time, making the system fragile and unreliable. The stochasticity and lack of transparency in LLM-based agents further exacerbate this issue, as their internal decision logic is not easily interpretable or verifiable [210,211]. Consequently, ensuring the predictability and controllability of emergent behavior remains a central challenge in designing safe and scalable Agentic AI systems.
4. **Scalability and Debugging Complexity:** As Agentic AI systems scale in both the number of agents and the diversity of specialized roles, maintaining system reliability and interpretability becomes increasingly complex [212,213]. This limitation stems from the *black-box chains of reasoning* characteristic of LLM-based agents. Each agent may process inputs through opaque internal logic, invoke external tools, and communicate with other agents all of which occur through multiple layers of prompt engineering, reasoning heuristics, and dynamic context handling. Tracing the root cause of a failure thus requires unwinding nested sequences of agent interactions, tool invocations, and memory updates, making debugging non-trivial and time-consuming.
   Another significant constraint is the system's *non-compositionality*. Unlike traditional modular systems, where adding components can enhance overall functionality, introducing additional agents in an Agentic AI architecture often increases cognitive load, noise, and coordination overhead. Poorly orchestrated agent networks where coordination, task delegation, and communication protocols are inadequately designed can result in redundant computation, contradictory decisions, or degraded task performance. Without robust frameworks for agent role definition, communication standards, and hierarchical planning, the scaling Agentic AI does not necessarily translate into greater intelligence or robustness. These limitations highlight the need for systematic architectural controls and traceability tools to support the development of reliable, large-scale agentic ecosystems.

5. **Trust, Explainability, and Verification:** Agentic AI systems pose huge challenges in explainability and verifiability due to their distributed, multi-agent architecture. While interpreting the behavior of a single LLM-powered agent is already non-trivial, this complexity is *multiplied* when multiple agents interact asynchronously through loosely defined communication protocols. Each agent may possess its own memory, task objective, and reasoning path, resulting in compounded opacity where tracing the causal chain of a final decision or failure becomes exceedingly difficult. The lack of shared, transparent logs or interpretable reasoning paths across agents makes it highly difficult, if not impossible, to determine *why* a particular sequence of actions occurred or which agent initiated a misstep. Compounding this opacity is the *absence of formal verification tools* tailored for Agentic AI. Unlike traditional software systems, where model checking and formal proofs offer bounded guarantees, there exists no widely adopted methodology to verify that a multi-agent LLM system comprising multiple large language model agents collaborating on tasks will perform reliably across all input distributions or operational contexts. This lack of verifiability presents a significant barrier to adoption in safety-critical domains such as autonomous vehicles, finance, and healthcare, where explainability and assurance are crucial. To advance Agentic AI safely, future research must address the fundamental gaps in causal traceability, agent accountability, and formal safety guarantees.
6. **Security and Adversarial Risks:** Agentic AI architectures introduce a significantly expanded attack surface compared to single-agent systems, exposing them to complex adversarial threats. One of the most critical vulnerabilities lies in the presence of a *single point of compromise*. Since Agentic AI systems are composed of interdependent agents communicating over shared memory or messaging protocols, the compromise of even one agent through prompt injection, model poisoning, or adversarial tool manipulation can propagate malicious outputs or corrupted state across the entire system. For example, a fact-checking agent fed with tampered data could unintentionally legitimize false claims, which are then integrated into downstream reasoning by summarization or decision-making agents.
Moreover, *inter-agent dynamics* themselves are susceptible to exploitation. Attackers can induce race conditions, deadlocks, or resource exhaustion by manipulating the coordination logic between agents. Without rigorous authentication, access control, and sandboxing mechanisms, malicious agents or corrupted tool responses can derail multi-agent workflows or cause erroneous escalation in task pipelines. These risks are amplified by the absence of standardized security frameworks for LLM-based multi-agent systems, leaving most current implementations defenseless against sophisticated multi-stage attacks. As Agentic AI moves toward broader adoption, especially in high-stakes environments, embedding secure-by-design principles and adversarial robustness becomes an urgent research priority.
7. **Ethical and Governance Challenges:** The distributed and autonomous nature of Agentic AI systems introduces huge ethical and governance concerns, particularly in terms of accountability, fairness, and value alignment. In multi-agent settings, *accountability gaps* emerge when multiple agents interact to produce an outcome, making it difficult to assign responsibility for errors or unintended consequences. This ambiguity complicates legal liability, regulatory compliance, and user trust, particularly in high-stakes domains such as autonomous vehicles, scientific research, or critical infrastructure management. Furthermore, *bias propagation and amplification* present a unique challenge: agents individually trained on biased data may reinforce each other's skewed decisions through interaction, leading to systemic inequities that are more pronounced than in isolated models. These emergent biases can be subtle and difficult to detect without ongoing monitoring over time or robust auditing mechanisms.
Additionally, *misalignment and value drift* pose serious risks in long-horizon or dynamic environments. Without a unified framework for shared value encoding, individual agents may interpret overarching objectives differently or optimize for local goals that diverge from human intent. Over time, this misalignment can lead to behavior that is inconsistent with ethical norms or user expectations. Current alignment methods, which are mostly designed for single-agent systems, are inadequate for managing value synchronization across heterogeneous agent collectives. These challenges highlight the urgent need for governance-aware agent architectures, incorporating principles such as role-based isolation, traceable decision logging, and participatory oversight mechanisms to ensure ethical integrity in autonomous multi-agent systems.
8. **Immature Foundations and Research Gaps:** Despite rapid progress and high-profile demonstrations, research and development in Agentic AI remains is still in early stage with unresolved issues that limit its scalability, reliability, and theoretical foundation. One of the central concerns is the *lack of standard architectures*. There is currently no widely accepted blueprint for how to design, monitor, or evaluate multi-agent systems built on LLMs . This architectural fragmentation makes it difficult to compare implementations, replicate experiments, or generalize findings across domains. Key aspects such as agent orchestration the structured coordination and role-based task allocation among agents along with memory structures and communication protocols, are often implemented in an ad hoc manner, leading to fragile systems that lack interoperability, consistency, and formal reliability guarantees.
Equally critical is the absence of causal foundations, as scalable causal discovery and reasoning remain unsolved challenges in current AI systems [214]. Causal discovery refers to the process of identifying underlying cause-and-effect relationships from data essential for understanding how different variables influence one another. Without the ability to represent and reason about these causal links, Agentic AI systems are inherently constrained in their ability to safely generalize beyond narrow, predefined training scenarios [186,215]. This limitation weakens their robustness when faced with distributional shifts, reduces their effectiveness in taking proactive actions, and impairs their ability to simulate alternative outcomes or hypothetical plans capabilities that are essential for intelligent coordination, adaptive planning, and high-stakes decision-making.
The gap between functional demos and principled design thus emphasizes an urgent need for foundational research in multi-agent system theory, causal inference integration, and benchmark development. Only by addressing these deficiencies can the field progress from prototype pipelines to trustworthy, general-purpose agentic frameworks suitable for deployment in high-stake environments.

### 5.1. Balanced critique and field-wide limitations

While AI Agents and Agentic AI systems offer significant promise, the field also faces notable limitations and unresolved critiques that merit careful attention. One central concern is the overreliance on LLMs, which, despite their generative capabilities, remain prone to hallucination, lack robust causal reasoning, and struggle with long-horizon planning. These limitations are compounded in Agentic AI systems, where emergent behavior, coordination complexity, and opaque reasoning chains can lead to unpredictable or unexplainable outputs. Moreover, current orchestration protocols often lack standardization, making agent interoperability and reproducibility difficult across platforms. From an ethical standpoint, persistent autonomy and memory retention raise concerns about surveillance, consent, and system accountability. Critics also highlight that many benchmark evaluations for agentic systems rely on artificial environments, failing to reflect the complexities of real-world deployment, especially in high-stakes domains like healthcare or finance. Additionally, the development of Agentic AI frameworks often emphasizes architectural novelty over rigorous empirical validation. These challenges underscore the need for not only technical innovation but also critical reflection, transparent evaluation metrics, and governance mechanisms. A balanced roadmap must address these critiques to ensure that agentic systems evolve with both functional robustness and ethical integrity.

## 6. Potential solutions and future roadmap

### 6.1. Potential solutions

To address the challenges and limitations of AI Agents and Agentic AI systems discussed in the previous section, we identify a set of promising solution pathways (as illustrated in Fig. 13) including RAG, tool-augmented reasoning, memory architectures, causal modeling, reflexive mechanisms, orchestration frameworks, and governance-aware designs. These techniques collectively represent the frontier of efforts to overcome the brittleness, scalability bottlenecks, and coordination challenges that currently constrain both AI Agents and Agentic AI. At present, most deployed systems rely heavily on heuristic wrappers, manual prompt engineering, and shallow coordination logic, falling short of robust autonomy and reliability. In the following several paragraphs, we examine how each solution targets specific technical or systemic limitations, highlight gaps in current implementations, and propose future research directions to evolve these solutions into mature, interoperable components of next-generation intelligent systems. This roadmap is essential for transitioning from ad hoc agent deployments to principled, generalizable frameworks capable of powering scalable, safe, and context-aware agentic ecosystems.

1. **RAG :** For AI Agents, RAG has the potential to mitigate hallucinations and can expand static LLM knowledge by grounding outputs in real-time data [216]. By embedding user queries and retrieving semantically relevant documents from vector databases like FAISS Source Link or Pinecone Pinecone, agents can generate contextually valid responses based on external facts. This retrieval-based grounding mechanism is particularly effective in domains such as enterprise search and customer support, where accuracy and access to up-to-date knowledge are essential for reliable task execution and user trust.

   In Agentic AI systems, RAG serves as a shared grounding mechanism across agents. For example, a summarizer agent may rely on the retriever agent to access the latest scientific papers before generating a synthesis. Persistent, queryable memory allows distributed agents to operate on a unified semantic layer, avoiding or minimizing inconsistencies due to divergent contextual views. When implemented across a multi-agent system, RAG helps maintain shared accuracy, enhances goal alignment, and reduces inter-agent misinformation propagation.

2. **Tool-Augmented Reasoning (Function Calling):** AI Agents benefit significantly from function calling, which extends their ability to interact with real-world systems [171,217]. Agents can query APIs, run local scripts, or access structured databases, thus transforming LLMs from static predictors into interactive problem-solvers [135,166]. This tool-augmented reasoning capability allows agents to dynamically access and process real-time, evolving information such as weather forecasts, stock prices, or user calendar updates and to perform executable actions like scheduling appointments, sending emails, or executing complex computations in Python. By bridging natural language reasoning with external tool interaction, this functionality empowers agents to go beyond static language generation and operate as autonomous, task-oriented decision-makers in real-world environments.

   For Agentic AI systems, function calling is instrumental in enhancing both autonomy and structured coordination among multiple agents. Each agent, assigned a specialized role within the system such as data retriever, visualizer, or decision-maker can independently invoke domain-specific APIs to perform targeted tasks, such as accessing clinical records or generating analytical dashboards. These function calls are not isolated; rather, they are embedded within an orchestrated pipeline a well-defined, multi-step workflow in which outputs from one agent seamlessly serve as inputs for the next [218]. This orchestration facilitates dynamic delegation, where agents can hand off subtasks based on predefined roles and capabilities without ambiguity or redundancy [22,26].

   Moreover, integrating function calling within such orchestrated pipelines establishes clearer behavioral boundaries between agents. Each agent operates within its defined scope of responsibility, reducing the likelihood of overlapping actions or conflicting decisions. When coupled with validation protocols (e.g., response verification or schema checks in Waitgpt [219]) and observation mechanisms (e.g., feedback loops or audit logs [220,221]), these boundaries are reinforced, ensuring that each agent not only performs its assigned task but does so transparently and accountably. This structured interaction model enhances system robustness, traceability, and ultimately the reliability of Agentic AI in complex, high-stakes domains.

3. **Agentic Loop: Reasoning, Action, Observation:** AI Agents often suffer from single-pass inference limitations [222]. The ReAct pattern introduces an iterative loop where agents reason about tasks, act by calling tools or APIs, and then observe results before continuing [136]. This feedback loop allows for more deliberate, context-sensitive behaviors. For example, an agent may verify retrieved data before drafting a summary, thereby reducing hallucination and logical errors. In Agentic AI, this pattern is critical for collaborative coherence. ReAct enables agents to evaluate dependencies dynamically reasoning over intermediate states, re-invoking tools if needed, and adjusting decisions as the environment evolves [136]. This loop becomes more complex in multi-agent settings where each agent's observation must be reconciled against others' outputs. Shared memory and consistent logging are essential here, ensuring that the reflective capacity of the system is not fragmented across agents.

4. **Memory Architectures (Episodic, Semantic, Vector):** As discussed before, AI Agents face limitations in long-horizon planning and session continuity. Memory architectures address this by persisting information across tasks [223]. Episodic memory allows agents to recall prior actions and feedback, semantic memory encodes structured domain knowledge, and vector memory enables similarity-based retrieval [224]. These elements are key for personalization and adaptive decision-making in repeated interactions. Agentic AI systems require even more sophisticated memory models due to distributed state management. Each agent may maintain local memory while accessing shared global memory to facilitate coordination. For example, a planner agent might use vector-based memory to recall prior workflows, while a QA agent references semantic memory for fact verification. Synchronizing memory access and updates across agents enhances consistency, enables context-aware communication, and supports long-horizon system-level planning.

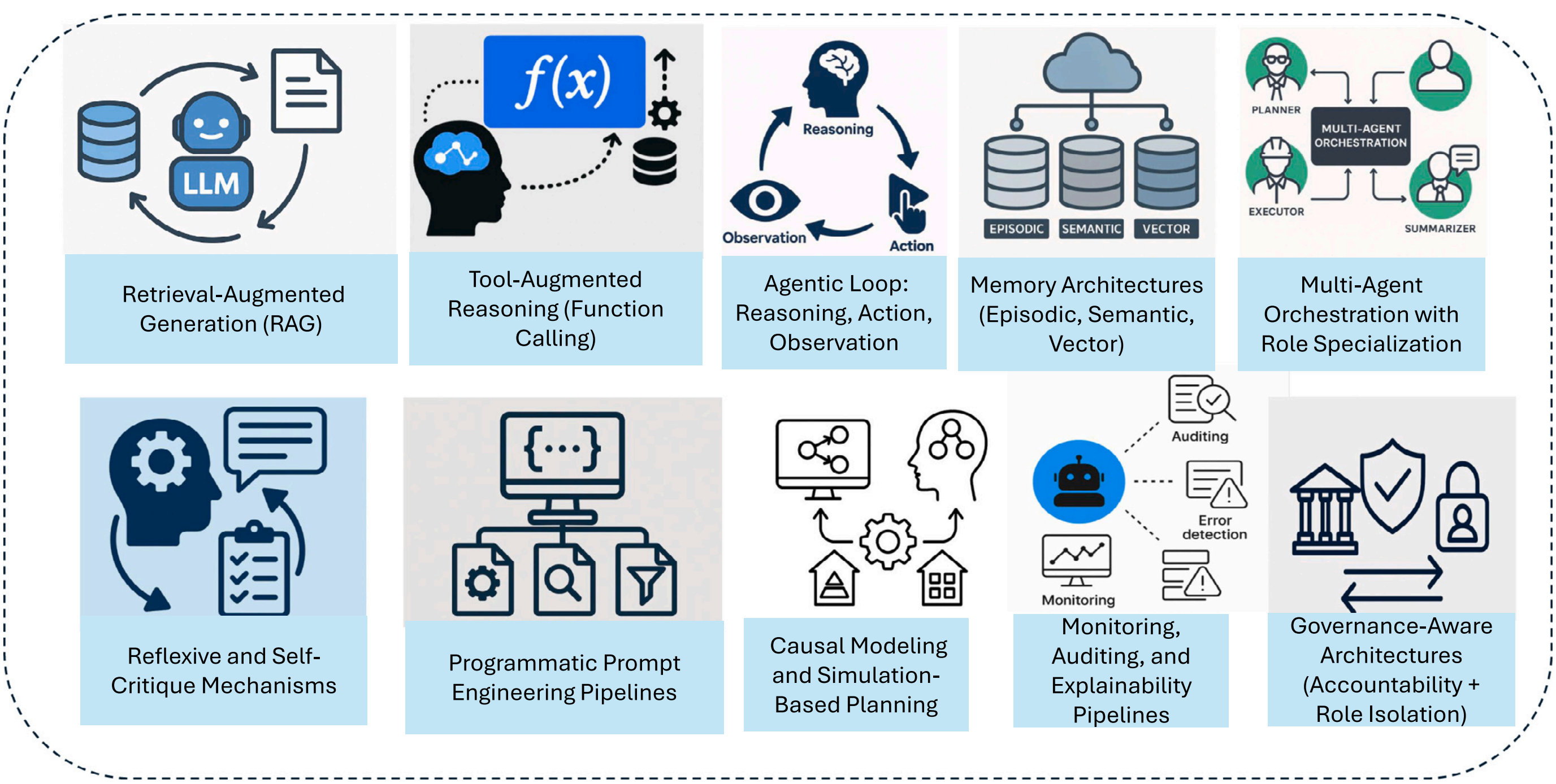

**Fig. 13.** Ten evolving architectural and algorithmic mechanisms such as RAG, tool augmentation, dynamic memory, causal modeling, orchestration, and reflexive self-evaluation are shown as key enablers to advance beyond prior usage toward addressing current limitations in reliability, scalability, and explainability. These techniques, while previously applied in isolated agent systems, are here recontextualized to support the demands of modern AI Agents and Agentic AI, enabling coordinated, adaptive, and verifiable behavior in increasingly complex and dynamic environments.

5. **Multi-Agent Orchestration with Role Specialization:** In conventional AI Agent systems, increasing task complexity is often addressed through modular prompt engineering or conditional branching logic. However, as the range and intricacy of tasks grow, a single agent may become overburdened, leading to performance degradation or failure to generalize effectively [225, 226]. To mitigate this, role specialization dividing the overall task into discrete functional units such as planning, summarization, or verification enables a form of compartmentalized reasoning even within a single-agent framework. In this context, compartmentalized reasoning refers to the simulation of distinct cognitive functions within one agent by prompting it to reason through subtasks in sequence, often mimicking multiple expert roles.
   In contrast, Agentic AI systems institutionalize orchestration as a core architectural feature. Here, orchestration refers to the dynamic coordination and task delegation across a team of specialized agents, each designed to handle a specific subfunction in the overall workflow. This is typically governed by a meta-agent or orchestrator, a supervisory agent responsible for allocating tasks, managing dependencies, and maintaining global context across all agents. Systems like MetaGPT and ChatDev exemplify this paradigm: agents adopt predefined professional roles such as CEO, software engineer, or reviewer and communicate through structured messaging protocols to collaboratively complete complex projects. This orchestrated, role-specialized design enhances system interpretability by isolating reasoning traces within clearly defined agent roles. It also improves scalability, as tasks can be parallelized across agents, and contributes to fault tolerance, as errors from one agent are contained and monitored by the orchestrator, preventing systemic failure. Such modular, coordinated architectures are foundational to building robust and transparent Agentic AI systems.
6. **Reflexive and Self-Critique Mechanisms:** AI Agents often fail silently or propagate errors. Reflexive mechanisms introduce the capacity for self-evaluation [227,228]. After completing a task, agents can critique their own outputs using a secondary reasoning pass, increasing robustness and reducing error rates. For example, a legal assistant agent might verify that its drafted clause matches prior case laws before submission. For Agentic AI, reflexivity extends beyond self-critique to inter-agent evaluation. Agents can review each other's outputs e.g., a verifier agent auditing a summarizer's work. Reflexion-like mechanisms ensure collaborative quality control and enhance trustworthiness [229]. Such patterns also support iterative improvement and adaptive replanning, particularly when integrated with memory logs or feedback queues [230,231].
7. **Programmatic Prompt Engineering Pipelines:** Manual prompt tuning introduces brittleness and reduces reproducibility in AI Agents. Programmatic pipelines automate this process using task templates, context fillers, and retrieval-augmented variables [232,233]. These dynamic prompts are structured based on task type, agent role, or user query, improving generalization and reducing failure modes associated with prompt variability. In Agentic AI, prompt pipelines enable scalable, role-consistent communication. Each agent type (e.g., planner, retriever, summarizer) can generate or consume structured prompts tailored to its function. By automating message formatting, dependency tracking, and semantic alignment, programmatic prompting prevents coordination drift and ensures consistent reasoning across diverse agents in real time [22,171].

8. **Causal Modeling and Simulation-Based Planning:** AI Agents often operate on statistical correlations rather than causal models, leading to poor generalization under distribution shifts. Embedding causal inference allows agents to distinguish between correlation and causation, simulate interventions, and plan counterfactually-informed, goal-directed actions that anticipate long-term effects and adapt to changing environments. For instance, in supply chain scenarios, a causally-aware agent can simulate the downstream impact of shipment delays. In Agentic AI, causal reasoning is vital for safe coordination and error recovery. Agents must anticipate how their actions impact others requiring causal graphs, simulation environments, or Bayesian inference layers. For example, a planning agent may simulate different strategies and communicate likely outcomes to others, fostering strategic alignment and avoiding unintended emergent behaviors. To enforce cooperative behavior, agents can be governed by a structured planning approach such as STRIPS or PDDL (Planning Domain Definition Language), where the environment is modeled with defined actions, preconditions, and effects. Inter-agent dependencies are encoded such that one agent's action enables another's, and a centralized or distributed planner ensures that all agents contribute to a shared goal. This unified framework supports strategic alignment, anticipatory planning, and minimizes unintended emergent behaviors in multi-agent systems.
9. **Monitoring, Auditing, and Explainability Pipelines:** AI Agents lack transparency, complicating debugging and trust. Logging systems that record prompts, tool calls, memory updates, and outputs enable post-hoc analysis and performance tuning. These records help developers trace faults, refine behavior, and ensure compliance with usage guidelines especially critical in enterprise or legal domains. Logging and explainability are even more critical for Agentic AI systems. With multiple agents interacting asynchronously, audit trails are essential for identifying which agent caused an error and under what conditions. Explainability pipelines that integrate across agents (e.g., timeline visualizations or dialogue replays) are key to ensuring safety, especially in regulatory or multi-stakeholder environments.
10. **Governance-Aware Architectures (Accountability and Role Isolation):** AI Agents currently lack built-in safeguards for ethical compliance or error attribution. Governance-aware designs introduce role-based access control, sandboxing, and identity resolution to ensure agents act within scope and their decisions can be audited or revoked. These structures reduce risks in sensitive applications such as healthcare or finance (For more applications refer to Table 10). In Agentic AI, governance must scale across roles, agents, and workflows. Role isolation prevents rogue agents from exceeding authority, while accountability mechanisms assign responsibility for decisions and trace causality across agents. Compliance protocols, ethical alignment checks, and agent authentication ensure safety in collaborative settings paving the way for trustworthy AI ecosystems.

### 6.2. Future roadmap

AI Agents are projected to evolve significantly through enhanced modular intelligence focused on five key domains as depicted in Fig. 14, which include proactive reasoning, tool integration, causal inference, continual learning, and trust-centric operations. The first transformative milestone involves transitioning from reactive to *Proactive Intelligence*, where agents initiate tasks based on learned patterns, contextual cues, or latent goals rather than awaiting explicit prompts. This advancement depends heavily on robust *Tool Integration*, enabling agents to dynamically interact with external systems, such as databases, APIs, or simulation environments, to fulfill complex user tasks. Equally critical is the development of *Causal Reasoning*, which will allow agents to move beyond statistical correlation, supporting inference of cause-and-effect relationships essential for tasks involving diagnosis, planning, or prediction. To maintain relevance over time, agents must adopt frameworks for *Continuous Learning*, incorporating feedback loops and episodic memory to adapt their behavior across sessions and environments. Lastly, to build user confidence, agents must prioritize *Trust & Safety* mechanisms through verifiable output logging, bias detection, and ethical guardrails especially as their autonomy increases. Together, these pathways will redefine AI Agents from static tools into adaptive cognitive systems capable of autonomous yet controllable operation in dynamic digital environments.

Agentic AI, as a natural extension of foundational AI agent frameworks, emphasizes collaborative intelligence through multi-agent coordination, contextual persistence, and domain-specific orchestration. Future systems (Fig. 14, right side) are expected to exhibit *Multi-Agent Scaling*, enabling specialized agents to operate concurrently under distributed control to tackle complex, high-dimensional problems mirroring collaborative workflows typical of human teams. This scaling necessitates a layer of *Unified Orchestration*, wherein orchestrators specialized meta-agents assume responsibility for dynamically assigning roles, managing inter-agent communication, sequencing task dependencies, and resolving potential conflicts. Orchestration, in this context, refers to the intelligent coordination and regulation of interactions among multiple autonomous agents to ensure coherent and efficient collective behavior. Sustained performance over time will depend on robust *Persistent Memory* architectures that allow agents to store and retrieve semantic, episodic, and shared task-relevant knowledge, supporting continuity in longitudinal operations and enabling agents to maintain awareness of evolving goals and environmental states. *Simulation Planning* will become a core capability, empowering agent collectives to model hypothetical decision trajectories, forecast consequences, and optimize courses of action through internal trial-and-error mechanisms thus reducing real-world risk and increasing adaptive robustness.

Moreover, establishing *Ethical Governance* frameworks will be crucial to ensure that agent collectives operate within aligned moral and legal boundaries. These frameworks will define accountability structures, verification mechanisms, and safety constraints, especially in high-stakes domains involving autonomous decisions. Finally, the emergence of *Domain-Specific Systems* tailored for sectors such as law, medicine, logistics, and climate science will allow Agentic AI to leverage contextual specialization and outperform general-purpose agents through fine-tuned workflows and expert reasoning capabilities.

A transformative direction for future AI systems is introduced by the *Absolute Zero: Reinforced Self-play Reasoning with Zero Data (AZR)* framework, which reimagines the learning paradigm for AI Agents and Agentic AI systems by removing dependency on external datasets [173]. Traditionally, both AI Agents and Agentic AI architectures have relied on human-annotated data, static knowledge corpora, or preconfigured environmental factors that constrain scalability and adaptability in open-world contexts. AZR addresses this limitation by enabling agents to autonomously generate, validate, and solve their own tasks, using verifiable feedback mechanisms (e.g., code execution) to ground learning. This self-evolving mechanism opens the door to truly autonomous reasoning agents capable of self-directed learning and adaptation in dynamic, data-scarce environments.

In the context of Agentic AI where multiple specialized agents collaborate within orchestrated workflows, meaning structured, coordinated processes managed by a central controller or meta-agent AZR lays the groundwork for agents to not only specialize in distinct roles but also co-evolve through self-improving interactions and shared learning objectives [173]. For instance, scientific research pipelines could consist of agents that propose hypotheses, run simulations, validate findings, and revise strategies entirely through self-play and verifiable reasoning, without continuous human oversight. By integrating the AZR paradigm, such systems can maintain persistent growth, knowledge refinement, and task flexibility over time. Ultimately, AZR highlights a future in which AI agents transition from static, pretrained tools to intelligent, self-evolving and -improving ecosystems positioning both AI Agents and Agentic AI at the forefront of next-generation artificial intelligence.

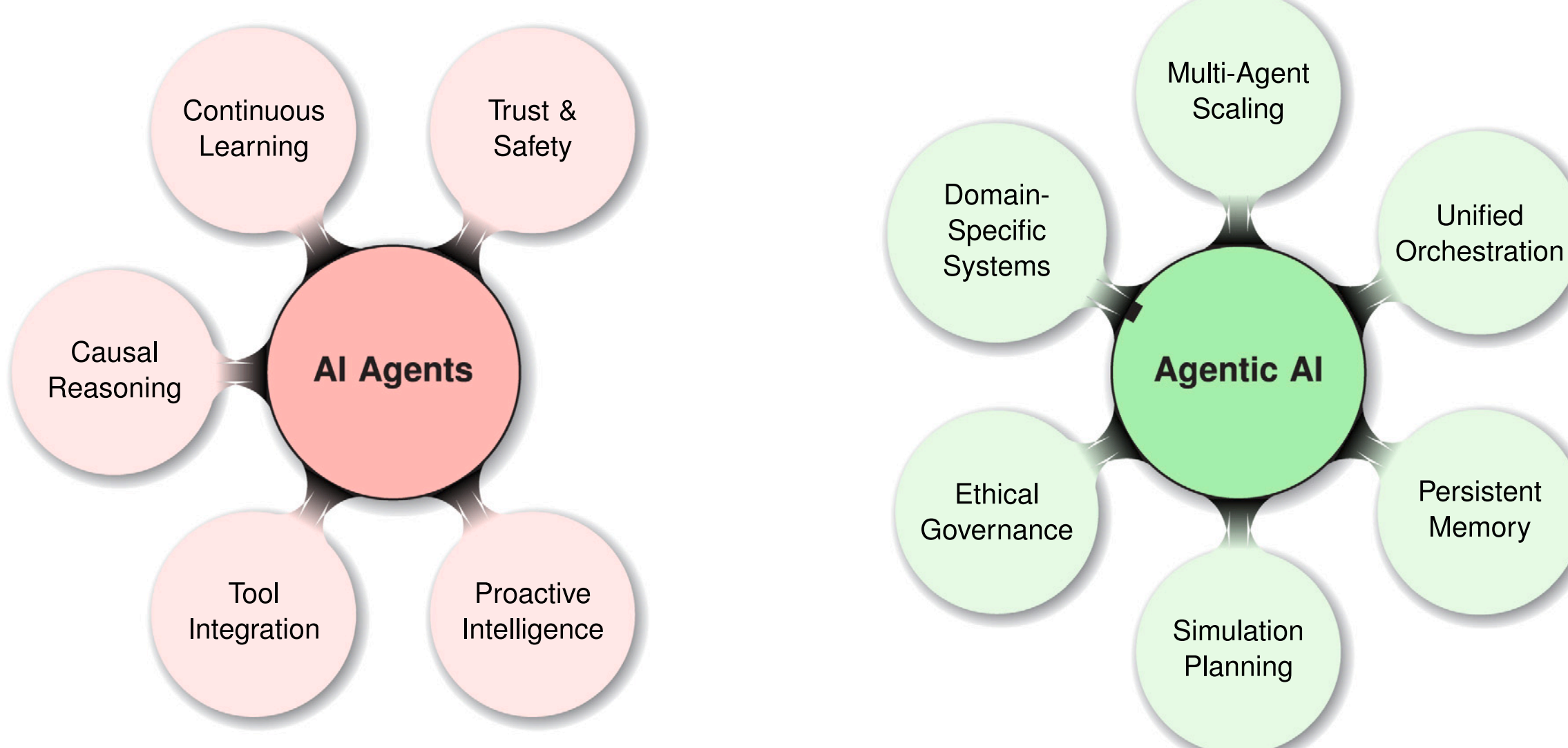

**Fig. 14.** Mind map contrasting future roadmaps of AI Agents (left) and Agentic AI (right). Each branch represents a unique trajectory, such as modular tool integration for AI Agents versus coordinated multi-agent planning, persistent memory, and adaptive orchestration for Agentic AI—highlighting divergent paths in autonomy and system design.

## 7. Conclusion

In this study, we presented a comprehensive literature-based evaluation of the evolving landscape of AI Agents and Agentic AI systems, offering a structured taxonomy that highlights foundational concepts, architectural evolution, application domains, and key limitations and potential solutions. Beginning with a foundational understanding, we characterized AI Agents as modular, task-specific entities with constrained autonomy and reactivity within the tasks specified for them. Their operational scope is enabled by the integration of LLMs and LIMs, which serve as core reasoning modules for perception, language understanding, and decision-making. We identified generative AI as a functional precursor to AI Agents, emphasizing its limitations in autonomy and goal persistence, and examined how LLMs drive the progression from passive generation to interactive task completion through tool augmentation.

This study then explored the conceptual emergence of Agentic AI systems as a transformative evolution from isolated agents or entities to orchestrated, multi-agent ecosystems that is, coordinated frameworks in which multiple specialized agents interact through structured role assignment, task delegation, and centralized or distributed control enabled by collaborative learning and collective decision making. We analyzed key differentiators such as distributed cognition, persistent memory, and coordinated planning that distinguish Agentic AI from conventional single-agent models. This analytical comparison was followed by a detailed breakdown of architectural evolution, highlighting the transition from monolithic, rule-based frameworks to modular, role-specialized networks. These networks are facilitated by orchestration layers, which serve as coordination mechanisms either centralized or decentralized that assign tasks, monitor agent interactions, and manage dependencies across specialized agents. Together with reflective memory architectures, these orchestration layers enable dynamic collaboration, task adaptability, and context preservation, marking a foundational shift toward scalable, intelligent agent collectives in Agentic AI systems.

Additionally, this study surveyed application domains in which these two paradigms (AI Agents and Agentic AI systems) are deployed. For AI Agents, we illustrated their role in automating customer support, internal enterprise search, email prioritization, and scheduling. For Agentic AI, we showcased use cases in collaborative research, swarm robotics, medical decision support, and adaptive workflow automation, supported by practical examples and industry-grade systems. Finally, this study provided a deep analysis of the challenges and limitations affecting both paradigms. For AI Agents, we discussed hallucinations, shallow reasoning, and planning constraints as the key challenges faced, while for Agentic AI, we addressed amplified causality issues, coordination bottlenecks, emergent behavior, and governance concerns limiting the rapid advancement and adoption of these systems.

Through this comparative framework, we conclude that AI Agents serve well in narrow, tool-integrated scenarios with defined goals, while Agentic AI represents a paradigm shift toward distributed, multi-agent cognition capable of autonomous planning and decision-making. The transition from reactive task execution to orchestrated, collaborative workflows marks a significant milestone in the evolution of intelligent systems. These insights offer a roadmap for the future development and deployment of trustworthy, scalable Agentic AI systems that are capable of adapting to complex real-world environments

## CRediT authorship contribution statement

**Ranjan Sapkota:** Writing – review & editing, Writing – original draft, Visualization, Methodology, Investigation, Formal analysis, Conceptualization. **Konstantinos I. Roumeliotis:** Writing – review & editing, Visualization, Methodology. **Manoj Karkee:** Writing – review & editing, Visualization, Supervision, Resources, Project administration, Investigation, Funding acquisition, Formal analysis.

## Statement on AI Writing Assistance

ChatGPT and Perplexity were utilized to enhance grammatical accuracy and refine sentence structure; all AI-generated revisions were thoroughly reviewed and edited for relevance. Additionally, ChatGPT-4o was employed to generate realistic visualizations.

## Declaration of competing interest

The authors declare that they have no known competing financial interests or personal relationships that could have appeared to influence the work reported in this paper.

## Acknowledgments

This work was supported in part by the National Science Foundation (NSF) and the United States Department of Agriculture (USDA), National Institute of Food and Agriculture (NIFA), through the "Artificial Intelligence (AI) Institute for Agriculture" program under Award Numbers AWD003473 and AWD004595, and USDA-NIFA Accession Number 1029004 for the project titled "Robotic Blossom Thinning with Soft Manipulators". Additional support was provided through USDA-NIFA Grant Number 2024-67022-41788, Accession Number 1031712, under the project "ExPanding UCF AI Research To Novel Agricultural Engineering Applications (PARTNER)". The partial financial support for open access publication was provided by the Hellenic Academic Libraries Link (HEAL-Link).

## Data availability

No data was used for the research described in the article.